\documentclass{article}

\PassOptionsToPackage{sort}{natbib}

\usepackage[final]{neurips_2025}

\newif\ifskipchecklist
\skipchecklisttrue

\usepackage[utf8]{inputenc} 
\usepackage[T1]{fontenc}    
\usepackage[english]{babel}
\usepackage[
    pdftex,
    pdfauthor={Benjamin Matthias Ruppik, Julius von Rohrscheidt, Carel van Niekerk, Michael Heck, Renato Vukovic, Shutong Feng, Hsien-chin Lin, Nurul Lubis, Bastian Rieck, Marcus Zibrowius, Milica Gasic},
    pdftitle={Less is More: Local Intrinsic Dimensions of Contextual Language Models}
]{hyperref}       
\usepackage{url}            
\usepackage{booktabs}       
\usepackage{amsfonts}       
\usepackage{nicefrac}       
\usepackage{microtype}      
\usepackage[dvipsnames]{xcolor} 
\usepackage{tikz}
\usepackage{pgfplots}
\usepackage{pgfplotstable}
\pgfplotsset{compat=1.7}
\usepgfplotslibrary{fillbetween} 
\usepackage{wrapfig}
\usepackage{titletoc}

\usepackage{multirow}
\usepackage[inline]{enumitem}

\addto\extrasenglish{  
    
}

\usepackage{siunitx} 

\usepackage{graphicx}   
\usepackage{subcaption} 

\usepackage{amsmath}
\usepackage{hyperref}
\usepackage[nameinlink]{cleveref} 

\usepackage[ruled,vlined]{algorithm2e} 

\RestyleAlgo{ruled} 
\SetKwComment{Comment}{/* }{ */}

\newif\ifdraft
\drafttrue                

\colorlet{citemarker}{RoyalBlue!60!black}

\newcommand{\MultiWOZ}{\texttt{MultiWOZ}}
\newcommand{\EmoWOZ}{\texttt{EmoWOZ}}
\newcommand{\SGD}{\texttt{SGD}}
\newcommand{\ICLR}{\texttt{ICLR 2024}}
\newcommand{\Reddit}{\texttt{Reddit}}
\newcommand{\Wikipedia}{\texttt{Wikipedia}}

\newcommand{\RoBERTa}{RoBERTa}
\newcommand{\GPTMedium}{GPT-2-medium}
\newcommand{\Phiminiinstruct}{Phi-3.5-mini-instruct}
\newcommand{\LlamaThreeEightB}{Llama-3.1-8B}

\newcommand{\EmbeddingModel}{\mathcal{M}}
\newcommand{\Tokenizer}{\mathcal{T}}
\newcommand{\EmbeddingModelWithIndex}{\EmbeddingModel_{i}}

\newcommand{\Datacorpus}{\mathcal{D}}

\newcommand{\DatacorpusSubsampleSize}{M}

\newcommand{\TokenVectorSubsampleSize}{N}

\newcommand{\LocalNeighborhoodSize}{L}

\newcommand{\Neighborhood}{\mathcal{N}}

\DeclareMathOperator{\localestimator}{dim}
\DeclareMathOperator{\twonn}{TwoNN} 
\newcommand{\twonntext}{TwoNN} 

\newcommand{\NN}{\mathbb{N}}
\newcommand{\RR}{\mathbb{R}}

\title{Less is More: Local Intrinsic Dimensions of Contextual Language Models}

\author{%
    Benjamin Matthias Ruppik{\(^{1}\)} ~ Julius von Rohrscheidt{\(^{2,3}\)} ~ Carel van Niekerk{\(^{1}\)} ~ Michael Heck{\(^{1}\)} \\
    \textbf{Renato Vukovic{\(^{1}\)} ~ Shutong Feng{\(^{1}\)} ~ Hsien-chin Lin{\(^{1}\)} ~ Nurul Lubis{\(^{1}\)}}   \\
    \textbf{Bastian Rieck{\(^{2,3,4}\)} ~ Marcus Zibrowius{\(^{1}\)} ~ Milica Ga\v{s}i\'c{\(^{1}\)}} \\
    \(^{1}\)\ Faculty of Mathematics and Natural Sciences, Heinrich Heine University Düsseldorf, Germany \\ 
    \(^{2}\)\ Institute of AI for Health, Helmholtz Munich, Germany \\
    \(^{3}\)\ Technical University of Munich, Germany \\
    \(^{4}\)\ University of Fribourg, Switzerland \\
    \texttt{\{ruppik,niekerk,heckmi,revuk100,fengs,} \\
    \texttt{linh,lubis,marcus.zibrowius,gasic\}@hhu.de} \\
    \texttt{julius.rohrscheidt@helmholtz-munich.de} \\
    \texttt{bastian.grossenbacher@unifr.ch}
}

\begin{document}

\maketitle

%
%
%
%
\begin{abstract}
Understanding the internal mechanisms of large language models (LLMs) remains a challenging and complex endeavor. 
Even fundamental questions, such as how fine-tuning affects model behavior, often require extensive empirical evaluation. 
In this paper, we introduce a novel perspective based on the geometric properties of contextual latent embeddings to study the effects of training and fine-tuning. 
To that end, we measure the local dimensions of a contextual language model's latent space and analyze their shifts during training and fine-tuning.
We show that the local dimensions provide insights into the model's training dynamics and generalization ability.
Specifically, the mean of the local dimensions predicts when the model’s training capabilities are exhausted, as exemplified in a dialogue state tracking task, overfitting, as demonstrated in an emotion recognition task, and grokking, as illustrated with an arithmetic task.
Furthermore, our experiments suggest a practical heuristic: reductions in the mean local dimension tend to accompany and predict subsequent performance gains.
Through this exploration, we aim to provide practitioners with a deeper understanding of the implications of fine-tuning on embedding spaces, facilitating informed decisions when configuring models for specific applications. 
The results of this work contribute to the ongoing discourse on the interpretability, adaptability, and generalizability of LLMs by bridging the gap between intrinsic model mechanisms and geometric properties in embeddings.
\end{abstract}

\section{Introduction}
\label{sec:introduction}

Large language models (LLMs) have transformed natural language processing in recent years, achieving impressive performance across a variety of tasks~\citep{devlin-etal-2019-bert,radford2018improving,brown2020language,touvron2023llama,jiang2023mistral}. 
These models learn contextual token embeddings in high-dimensional latent spaces, whose structure governs how information is represented and processed. 
However, most performance diagnostics in LLMs rely on supervised validation or task-specific probes, with few attempting to understand the geometry of the LLM embedding spaces. 
%
%
Our paper is motivated by a central question: \emph{Can structural changes in the embedding space yield unsupervised insights into model behavior across language modeling tasks?} 
%
We address this question by applying a localized version of the {\twonntext} estimator \citep{facco2017estimating} to quantify the \emph{local intrinsic dimension} of contextual token embeddings. 
While the ambient dimension of these embeddings is typically large (ranging from hundreds to thousands of dimensions), our observations show that the local intrinsic dimension reflects a lower-dimensional manifold structure that varies across data space regions. 
This heterogeneity allows us to derive model- and data-specific geometric signatures without relying on labeled validation data or task-specific supervision.

\textbf{Contributions} 
We present a unified framework for analyzing LLM training dynamics through dimensionality.  
Based on this framework, we
\begin{enumerate*}[label=(\roman*)]
    \item demonstrate that fine-tuning reshapes the local intrinsic dimension in a \emph{dataset-specific} manner, allowing us to infer overlap between training data used for fine-tuning with validation (or test) data from geometric shifts alone, without requiring labels;
    \item show that dimensionality anticipates the onset of \emph{grokking} in synthetic arithmetic tasks, identifying generalization beyond training data from the dimensionality of the model's embeddings of the training data alone;
    \item demonstrate that local intrinsic dimensions can indicate \emph{training convergence}, as evidenced in a sequence-tagging-based dialogue state tracking task where stabilizing dimension correlates with model stabilization; and
    \item show that local intrinsic dimensions detect \emph{overfitting} in a sequence classification setting, where an initial drop followed by a rise in dimension reflects the model’s tradeoff between generalization and memorization.
\end{enumerate*}
In total, our results suggest that local intrinsic dimensions serve as a valuable unsupervised signal for practitioners seeking to interpret and monitor LLM behavior. 
Beyond offering empirical insights, our work thus highlights the potential of geometric descriptors to complement traditional evaluation methods and potentially serve to inform future model design.\footnote{Our code is available at \url{https://github.com/aidos-lab/Topo_LLM_public} and \url{https://github.com/aidos-lab/grokking-via-lid}.}

\section{Related Work}


Our work lies at the intersection of geometry-aware language model analysis and dataset-level representation diagnostics. 
While prior efforts focus on the intrinsic dimensions of single sequences or model parameters, our contribution is quantifying local intrinsic dimensions across datasets and training stages, enabling unsupervised inspection of generalization dynamics and fine-tuning effects.

\paragraph{Notions of intrinsic dimensions of language models} 
In a first qualitative analysis of the internal representations of transformer models, \citet{ethayarajh-2019-contextual} and \citet{cai2021isotropy} identify clusters and low-dimensional manifold structures in contextual embedding spaces.
Recent work has started studying relationships between geometric properties of LLMs and corresponding semantics.
In \cite{tulchinskii2024intrinsic}, the authors discover that the average intrinsic global dimension of artificially generated texts is lower than that of human-written texts.
In that work, the points in the latent space are taken from a single text paragraph, so their work applies to single data input sequences, not to entire datasets like our approach.
\cite{valeriani2024geometry} investigate how the global intrinsic dimension estimated is altered as data is passed through an LLM.
\cite{aghajanyan2020intrinsic} define a notion of intrinsic dimension based on restricting the model's parameter space and its effect on the objective function, and show that larger LLMs tend to have smaller intrinsic dimensionality.
That work studies the dimensionality of the model's parameter space, and not the latent space created from specific datasets.
In \cite{viswanathan2025geometrytokensinternalrepresentations}, the authors analyze token-level intrinsic dimensions, linking token geometry to model next-token-prediction loss.
The difference between their method and ours is that they always consider the contextual token embeddings of a single, sufficiently long prompt as making up the embedding space. 
In contrast, we sub-sample from an entire dataset split. 
\cite{lee2025shared} identify structural similarities between the token embedding spaces of different language models, particularly reporting correlations in local intrinsic dimension measures. 
However, their analysis is limited to the non-contextual token embedding and unembedding layers, without considering contextual representations.

\paragraph{Topology-based analysis of models}
The topological descriptors in \cite{kushnareva-etal-2021-artificial,perez2022topological,tulchinskii2024intrinsic,lee-etal-2025-geometric} correspond to entire sequences of input data, while our proposed estimates are defined on the token level.
In \cite{durrani-etal-2022-transformation}, pre-trained and fine-tuned models are compared using hierarchical clustering and alignment functions. 
The results indicate that the latent space in the higher layers adapts to task-specific concepts, while the lower layers preserve the general concepts learned in the pre-trained model.
\cite{eddib2024gelorageometricadaptiveranks} introduces a method for fine-tuning large language models by dynamically adjusting LoRA ranks based on the intrinsic dimensionality of hidden state representations.
Here, the intrinsic dimension is measured as the rank of an information matrix, and not from the data space as in our work.
In \citet{chang-etal-2022-geometry}, the representational geometry of multilingual language models is explored, focusing on how they balance encoding language-specific and language-agnostic features within a shared multilingual space.
\citet{ruppik-etal-2024-local} utilize topological features derived from neighborhoods in a contextual embedding space to improve performance in a sequence tagging task. 
The definition of their features requires a fixed ambient data corpus, whereas our methods give an intrinsic measure of the dataset under consideration; thus, our measures are comparable between different model checkpoints of the feature generation model. 
Both works are situated in the nascent field of \emph{topological deep learning}~\citep{Papamarkou24a}, employing concepts from geometry and topology in an \emph{observational manner} to study machine learning models.
In contrast to mechanistic interpretability (see, for example, the survey by \citealt{ferrando2024primerinnerworkingstransformerbased}), in which one tries to explain how a language model's internal representations fit together into higher-level abstractions, our approach is more atomic.
We study the smallest non-divisible representation in the model (the contextual embedding of single tokens) and the space created by collections of these.

\section{Methods}
\label{sec:methods}

\subsection{Large Language Models}
\label{sec:large_language_models}

Modern contextual language models are typically either masked language models (MLMs) \citep{devlin-etal-2019-bert,liu2019roberta} or autoregressive language models (ALMs) \citep{radford2019language}. 
MLMs predict masked tokens based on bidirectional context, while ALMs generate tokens sequentially, conditioned on preceding tokens. 
In both cases, the representation of a token in context is given by a vector at each model layer, enabling geometric analysis of a data corpus on a layer-wise basis. 
We now give an overview of this construction; the whole procedure is summarized in \autoref{alg:compute_local_estimates}.

\subsection{Latent Space Modeling}
\label{sec:laten_space_modeling}

Let $\Datacorpus = (s_0, s_1, \ldots, s_D)$ be a text corpus and $\EmbeddingModel$ a language model (MLM or ALM) of depth $l$ with tokenizer $\Tokenizer$. Each sequence \(s_m\) yields tokenized input:
\begin{equation}
    \Tokenizer(s_m) = \left(t_0^m, \ldots, t_{n_m}^m\right),
\end{equation}
with corresponding contextualized token embeddings at layer \(i\):
\begin{equation}
    \EmbeddingModelWithIndex(s_m) = \left(\EmbeddingModelWithIndex(t_0^m), \ldots, \EmbeddingModelWithIndex(t_{n_m}^m)\right).
\end{equation}
We distinguish between \emph{regular} embeddings (tokens embedded as-is during a forward-pass) and \emph{masked} embeddings (specific tokens replaced by \texttt{[MASK]}, requiring the model to infer their representations from the surrounding context). Despite the simplified notation, embeddings are context-sensitive.
The token embedding point cloud at layer \(i\) is:
\begin{equation}
    \mathbb{T}_i = \{ \EmbeddingModelWithIndex(t_j^m) \}_{m = 0, \ldots, D;\ j \in I_m},
\end{equation}
with \(I_m\) denoting token indices in the \(m\)-th sequence.
Distances between points are measured in the ambient Euclidean space.
In practice, $\mathbb{T}_i$ can contain millions of vectors, making subsequent neighborhood computation on the full dataset infeasible.

To obtain a representative subset of tokens $\mathbb{T}$, we take a two-step sampling approach:
We first sample \(\DatacorpusSubsampleSize\) sequences from $\Datacorpus$, and after de-duplication, we sample \(\TokenVectorSubsampleSize\) vectors from the resulting token embedding space. 
Finally, we compute neighborhoods \(\Neighborhood_{L}(t_j; \mathbb{T})\) for each token using a locality parameter \(L\). 
We assume the dataset is drawn from an underlying text-generating distribution, and our sampling approximates its geometric structure. 
We confirm this assumption in our sensitivity analysis in \Cref{sec:sensitivity_analysis}.

\begin{algorithm}[htb]
    \LinesNumbered
    \caption{
        \label{alg:compute_local_estimates}
        \textsc{Compute Local Dimension Estimates}
    }
    \SetKwInput{Input}{Input~}
    \SetKwInput{Output}{Output~}
    \BlankLine
    \Input{\parbox[t]{\textwidth}{%
        Text corpus \(\Datacorpus\);
        \newline
        Embedding model \(\EmbeddingModel\) (with corresponding tokenizer \(\Tokenizer\));
        \newline
        Dimension estimator \(\localestimator\) (default \(\localestimator = \twonn\));
        \newline
        Three parameters:\\
        – Size of text corpus sequence sub-sample \(\DatacorpusSubsampleSize \in \NN\);\\
        – Size of token sub-sample \(\TokenVectorSubsampleSize \in \NN\);\\
        – Local neighborhood size \(\LocalNeighborhoodSize \in \NN\).
        }
    }
    \Output{Token-level local estimates given as a vector in \(\RR^{\TokenVectorSubsampleSize}_{\ge 0}\), i.e., one non-negative number for each of the sub-sampled tokens.}
    \BlankLine
    Take a random sub-sample \(\Datacorpus_{\sim \DatacorpusSubsampleSize}\) of size \(\DatacorpusSubsampleSize\) of the dataset sequences \;
    Compute embeddings for each token in each sequence of \(\Datacorpus_{\sim \DatacorpusSubsampleSize}\) \;
    De-duplicate the embedding vectors \;
    Take token vector sub-sample of size \(\TokenVectorSubsampleSize\), call the resulting set of token vectors \(\mathbb{T}\) \;
    \For{all $t_{j} \in \mathbb{T}$}{
        Compute the \(L\) nearest neighbors of \(t_{j}\) in \(\mathbb{T}\), resulting in the neighborhood \(\Neighborhood_{L}(t_{j}; \mathbb{T})\) \;
        Compute the local estimate \(\localestimator(\Neighborhood_{L}(t_{j}; \mathbb{T})) \in \RR_{\ge 0}\) of this neighborhood \;
    }
    \Return{The vector of token-level dimension estimates.}
\end{algorithm}

\subsection{Comparing Latent Spaces Across Models}

We compare a base model \(\EmbeddingModel\) with its fine-tuned variant \(\EmbeddingModel^{\Delta}\), trained on a corpus \(\Delta\). Since both models share the same architecture and tokenizer, there exists a canonical bijection between \(\EmbeddingModelWithIndex(\Datacorpus)\) and \(\EmbeddingModelWithIndex^{\Delta}(\Datacorpus)\), mapping each token’s representation in the base model to its counterpart in the fine-tuned model. More generally, such a bijection arises whenever the underlying model architectures and tokenizers are identical. After identical sub-sampling, this allows for point-wise comparison of the resulting latent spaces to evaluate geometric changes. 
Next, we define the local estimates used for this comparison.

\subsection{Local Dimension Estimates}
\label{sec:local_estimates}

The {\twonntext} estimator \citep{facco2017estimating} approximates the intrinsic dimension of a point cloud as a positive real number using only the distances \(r_1\) and \(r_2\) of each point to its nearest and second-nearest neighbors. 
It turns out that under weak assumptions, the ratio $\frac{r_2}{r_1}$ of these distances is Pareto-distributed~(see \citet{denti2021distributional} for details).
Then, the intrinsic dimension can be estimated from parameters describing this distribution, as long as the density around the points is locally approximately constant on the scale defined by the distance of the second nearest neighbor. 

We turn this into a \emph{local measure} as follows:
Let \(X\) be a point cloud and \(v \in X\). 
For \(\LocalNeighborhoodSize \in \NN_{>0}\), we define the \(\LocalNeighborhoodSize\)-local {\twonntext} estimate of \(X\) at \(v\) as \(\twonn(\Neighborhood_{\LocalNeighborhoodSize}(v; X))\), i.e., as the {\twonntext} estimate of the local neighborhood \(\Neighborhood_{\LocalNeighborhoodSize}(v; X)\) of \(v\) of size \(\LocalNeighborhoodSize\) in \(X\).
In this way, the parameter $L$ controls the locality scale of the dimension estimation.
In the context of our model \(\EmbeddingModel\) and corpus \(\Datacorpus\), we take \(X = \mathbb{T}\) and compute a vector of positive real-valued numbers \(\twonn(\Neighborhood_{\LocalNeighborhoodSize}(v; X))_{v \in \mathbb{T}} \in \RR^{\TokenVectorSubsampleSize}_{\ge 0}\), one for each token \(v \in \mathbb{T}\). 
We subsequently aggregate this to a mean estimate per corpus/model setup.
While sensitive to hyperparameters such as \(L\) and sample sizes, we show in \Cref{sec:sensitivity_analysis} that estimates remain stable under reasonable settings, enabling consistent comparisons across datasets and models.

\section{Experiments}
\label{sec:experiments}

We now apply our method to analyze how embedding spaces evolve during LLM-related learning tasks. 
Unless stated otherwise, all results pertain to the model’s final hidden layer, i.e., $\mathbb{T}_{-1}$ as defined in \Cref{sec:methods}. 
We refer the reader to \Cref{sec:layerwise_computation} for results on layers other than the last.
Subsequently, we will focus on four central questions:
\begin{enumerate}[label={(Q\arabic*)}, nosep]
    \item How does fine-tuning on different datasets alter latent space geometry?
    \item How can local dimension estimates detect \emph{grokking}?
    \item How can local dimension estimates detect the limit of \emph{training capabilities}?
    \item How can local dimension estimates detect \emph{overfitting}?
\end{enumerate}

\subsection{Fine-Tuning Induces Dataset-Specific Shifts in Heterogeneous Local Dimensions}
\label{sec:local_dimensions_language_modeling_finetuning}

To understand how fine-tuning alters model representations, we investigate the distribution of local intrinsic dimensions across token embeddings. 
These dimensions, estimated using the method introduced in \Cref{sec:methods}, reveal nuanced changes in the geometry of the embedding space.

\begin{wrapfigure}{O}{0.46\linewidth}
    \vspace{-1.4\baselineskip}
    \centering

    %
    \begin{subfigure}[t]{0.94\linewidth}
        \centering
        \includegraphics[width=\linewidth]{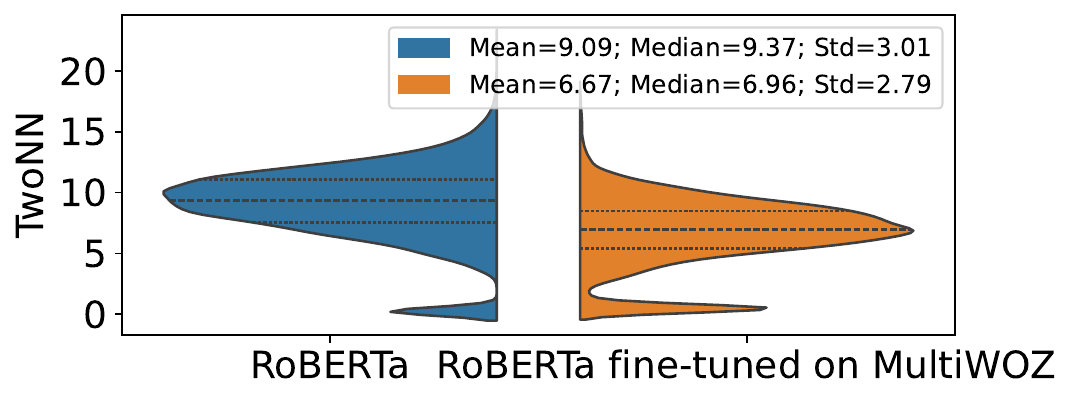}
        \caption{{\twonntext} estimates on {\MultiWOZ} validation}
    \end{subfigure}
    
    \vspace{0.5\baselineskip}
    
    \begin{subfigure}[t]{0.94\linewidth}
        \centering
        \includegraphics[width=\linewidth]{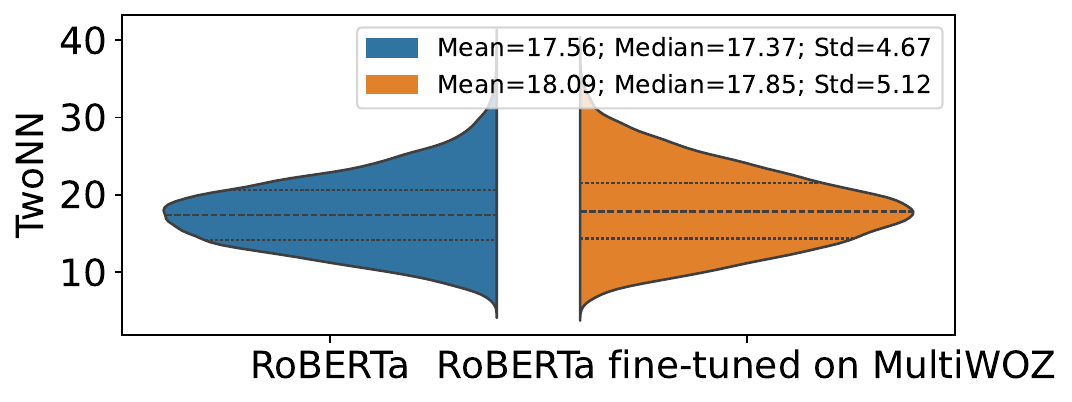}
        \caption{
            {\twonntext} estimates on {\Wikipedia} validation
        }
    \end{subfigure}
    
    \vspace{0.5\baselineskip}
    
    \begin{subfigure}[t]{0.94\linewidth}
        \centering
        \includegraphics[width=\linewidth]{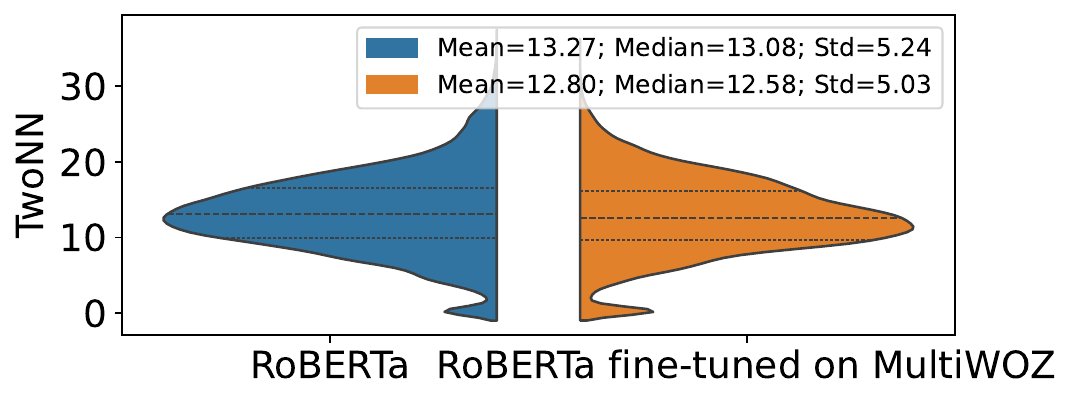}
        \caption{{\twonntext} estimates on {\Reddit} validation}
    \end{subfigure}

    \caption[Intrinsic‐dimension comparison for three data sources]{%
        Comparison of local intrinsic dimensions (LIDs) across three data modalities.
        The distribution of the local estimates over tokens is shown in the violin plot, together with their means and quartiles.
        The LID of embeddings originating from the fine‑tuning distribution ({\MultiWOZ}) differs markedly between models, whereas the LIDs for the out‑of‑distribution corpora ({\Wikipedia}, {\Reddit}) are almost indistinguishable.
        \label{fig:violin_main}
    }
    \vspace{-1.4\baselineskip} 
\end{wrapfigure}

\paragraph{Setup}

We evaluate models on datasets with varying domains and familiarity to the base and fine-tuned models: 
{\MultiWOZ}\texttt{2.1} \citep{eric-etal-2020-multiwoz}: Human-human multi-domain dialogues;
\texttt{Schema-Guided Dialogue} {\SGD} \citep{rastogi2020towards}: Human–virtual assistant dialogues; 
{\Reddit}: Reddit comments from the year 2022 mentioning Tesla, Inc.;
{\ICLR} Submissions: Titles and abstracts of ICLR 2024 papers collected by us; 
and {\Wikipedia}: The Hugging Face \texttt{wikitext-103-v1} corpus.

As can be seen from the selection of pre-training datasets of the respective models \citep{liu2019roberta,radford2019language}, this choice allows us to compare: 
(i) distributions not seen during pre-training or fine-tuning (e.g., {\Reddit}, {\ICLR}, which were released after the pre-training of the model concluded), 
(ii) distributions seen during pre-training ({\Wikipedia}), and 
(iii) distributions used for fine-tuning (e.g., {\MultiWOZ}, {\SGD}).
This contrast allows us to probe how local intrinsic dimensions behave in both seen and unseen data regimes.
We split {\Reddit} and {\Wikipedia} into training (80\%), validation (10\%), and test (10\%) subsets. 
Chronologically-ordered datasets (e.g., {\Reddit}) are shuffled before splitting to avoid temporal bias. 
\texttt{Wikipedia} was pre-processed by removing headings, empty lines, and stripping leading and trailing whitespace. 
Further dataset statistics are available in \Cref{tab:dataset_sizes}. 

Fine-tuning of the {\RoBERTa}-base models \citep{liu2019roberta} is performed using masked language modeling with a masking probability of \(0.15\). 
Each model is trained for 5 epochs on \num{10000} training examples using a batch size of 8, a learning rate peaking at \(5 \cdot 10^{-5}\) with 500 warmup steps, and linear decay thereafter. 
Weight decay of \(0.01\) is applied throughout.
For evaluation, we embed a subset of the \emph{validation} split of each dataset using both the base and fine-tuned model, ensuring the embeddings are out-of-sample relative to the fine-tuning set.

Our discussion in this section focuses on the masked language model {\RoBERTa}, whose architectures remain widely used in practice where encoder-based models are needed \citep{warner-etal-2025-smarter} and autoregressive alternatives are less suitable, such as in information retrieval.
Results for additional fine-tuning of the autoregressive models {\GPTMedium}, {\Phiminiinstruct}, and {\LlamaThreeEightB} can be found in \Cref{appendix:additional_finetuning}, where similar observations hold.

\paragraph{Results}

We find that local intrinsic dimensions vary significantly across tokens, as illustrated by the wide spread of the {\twonntext} distributions in \Cref{fig:violin_main}. 
Here, we choose the parameters \(\DatacorpusSubsampleSize=\num{7000}\), \(\TokenVectorSubsampleSize=\num{60000}\), \(\LocalNeighborhoodSize=\num{128}\). 
This heterogeneity is consistent across different models and datasets. 
It aligns with prior observations that numerical data is often not confined to a single manifold of uniform dimensionality, but is instead composed of multiple regions with varying local geometry \citep{brown2023verifying}. 
These findings reinforce our decision to favor local over global estimates of embedding dimensionality.

Our central observation is that fine-tuning systematically lowers the local intrinsic dimension \emph{only} on the dataset used during the fine-tuning process. 
As shown in \Cref{fig:violin_main}a, the {\twonntext} estimates for the {\MultiWOZ}-validation embeddings are markedly lower for the fine-tuned {\RoBERTa} model than for the base model. 
By contrast, embeddings of unrelated datasets—such as {\Wikipedia} (\Cref{fig:violin_main}b) or {\Reddit} comments (\Cref{fig:violin_main}c)—show no or only minimal changes in their local dimension distributions. 
Quantitatively, this observation can be validated by the standardized mean difference between the respective cohorts, with a value of $1.19$ for {\MultiWOZ}, and $0.08$ (resp. $0.1$) for {\Reddit} (resp. {\Wikipedia}).

We find this shift to be stable across training, validation, and test splits of the same dataset, suggesting that it reflects robust structural changes in the latent space rather than artifacts (see \Cref{sec:sensitivity_analysis} for a thorough sensitivity analysis showing the method's robustness to the choice of sampling parameters). 
This behavior highlights strong dataset-specificity in how local geometry adapts under fine-tuning. 
The effect is absent when evaluating on out-of-distribution data, supporting the view that local dimension reductions correspond to improved fit and specialization on the fine-tuned task, while leaving unrelated regions of the embedding space unchanged.

\subsection{Local Dimensions Detect Grokking}
\label{sec:local_dimensions_grokking}

\emph{Grokking} is the phenomenon that a machine learning model acquires specific skills only after extended training far beyond overfitting on the training set, which was first discovered by \citet{power2022grokking}.
In such cases, from the model's performance on the training set alone, it is a complex problem to predict under which choices of hyperparameters grokking will occur \citep{junior2024predicting}.

\paragraph{Setup}

We here consider the task of learning an arithmetic operation on a small group (addition modulo \(p\)).
For example, for \(p=197\) such an input sequence would be \texttt{[155, `o', 88, `=']}, where the model should predict the result \texttt{46}.
The tokenization is constructed so that every operand in the group is encoded as a single token, with the addition of the operation token \texttt{`o'} and the equality token \texttt{`='}.
A certain fraction of all valid expressions in the group, ranging from 10 percent to 50 percent, is taken as the training set, and the remainder will be taken as the validation set.

We use a tiny decoder-only transformer model with two layers, 128 hidden dimensions, and four attention heads trained from scratch.
Optimization is performed by AdamW \citep{loshchilov2017decoupled} with learning rate \(0.001\) (with linear schedule over 400k steps, warmup of 10 steps), batch size 512, weight decay \(0.01\). 
Models are trained for many steps beyond the point where training performance saturates, keeping the same setup as in the original paper \citep{power2022grokking}.

For the local estimates, we set the number of sampled tokens \(\TokenVectorSubsampleSize\) as the minimum of 3000 and the maximum number of tokens available, and a neighborhood size \(\LocalNeighborhoodSize\) of 64.
Since the number of input sequences is comparatively small, we choose \(\DatacorpusSubsampleSize\) to encompass everything, and only subsample in the token selection step.
Still, we observe a similar qualitative behavior for other reasonable choices of these hyperparameters.
The hidden states at the last layer are sampled from all available tokens, which could come from the operands and the operation or equality symbol.

%
\begin{figure}[t]
    \centering
    \begin{subfigure}{0.325\linewidth}
        \centering
        \includegraphics[width=1.0\linewidth]{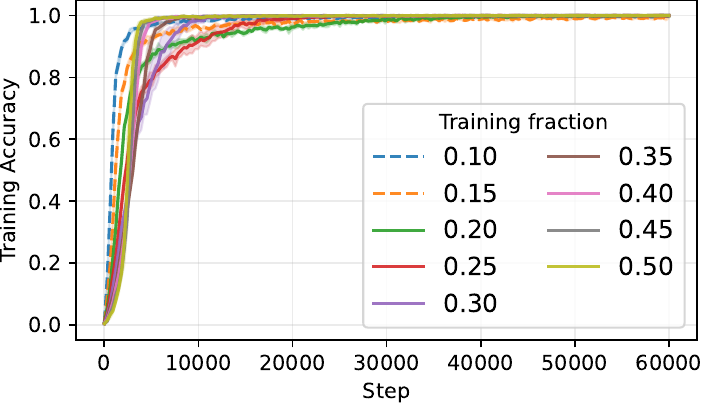}
        \caption{
            \textbf{Training} accuracy
            \label{fig:grokking_train.accuracy_grouped_by_frac_train}
        }
    \end{subfigure}
    \hfill
    \begin{subfigure}{0.325\linewidth}
        \centering
        \includegraphics[width=1.0\linewidth]{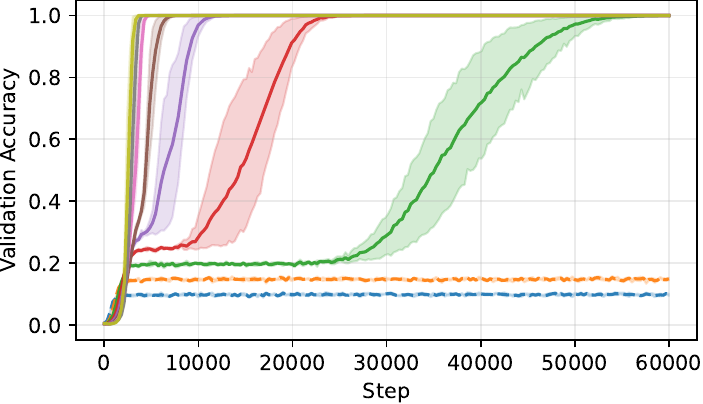}
        \caption{
            Validation accuracy
            \label{fig:grokking_val.accuracy_grouped_by_frac_train}
        }
    \end{subfigure}
    \hfill
    \begin{subfigure}{0.325\linewidth}
        \centering
        \includegraphics[width=1.0\linewidth]{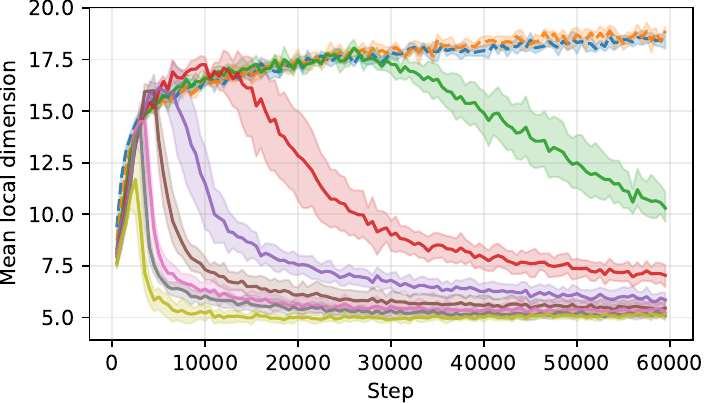}
        \caption{
            \textbf{Training} mean local estimates
            \label{fig:grokking_train.take_all.samples=3000.n-neighbors=64.mean_grouped_by_frac_train}
        }
    \end{subfigure}
    \caption{
        Training a model on addition mod \(p = 197\) with different training data fraction selected from \(\{0.1; 0.15; 0.2; 0.25; 0.3; 0.4; 0.5\}\).
        The plots show the development for \num{60000} batches, with mean and 95\% confidence interval over \(5\) training seeds per configuration (plots per seed are in \Cref{appendix:additional_grokking}).
        The mean local estimates are computed on the \textbf{training split} for the parameters \(\TokenVectorSubsampleSize=3000\); \(\LocalNeighborhoodSize=64\).
        Dashed lines highlight the runs where grokking did \emph{not} occur.
        \label{fig:grokking_p_197_grouped_by_frac_train}
    }
\end{figure}

\paragraph{Results}

Performance measures versus local estimates are shown in \Cref{fig:grokking_p_197_grouped_by_frac_train} for the first \num{60000} training steps averaged over seeds (we present plots for individual seeds in \Cref{fig:grokking_p_197_individual_seeds} in the appendix).
The training accuracy quickly reaches an almost perfect score in all cases, with the validation accuracy lagging behind.
In the time frame under consideration, the model can generalize and reach almost perfect validation accuracy only in those runs for which the training data portion exceeds 20\%. 
Based on the loss and accuracy of the training data alone, one would not be able to predict which configurations can break out and generalize to the validation data.

A typical pattern for the mean local dimension computed on the training data in all these runs is that it increases in the first few thousand global steps~(see \Cref{fig:grokking_train.take_all.samples=3000.n-neighbors=64.mean_grouped_by_frac_train}).
But subsequently, the training mean local dimension starts dropping significantly for those runs that exhibit grokking.
Observe that the timing of this drop coincides with the start of the increasing validation accuracy in \Cref{fig:grokking_val.accuracy_grouped_by_frac_train}.
We can conclude that in this setting, a drop in the mean local dimension on the training set strongly indicates a successful generalization to unseen validation data.

For those runs in which the validation accuracy does not increase beyond the fraction of training data during the selected training time (with 10\% and 15\% of training data, highlighted via dashed lines), the local dimension increases and then stays mostly flat.
Quantitatively, the Spearman rank correlation between the training/validation accuracy and mean local dimension measured over the training split is positive for those training fractions where the model does not grok (\(0.880/0.076\) and \(0.922/0.257\) for training fractions 10\% and 15\%).
This behavior suggests that the model only tries to learn the training examples by heart and fails to generalize to unseen data.
For all runs grouped by training fraction \(\ge 20\)\%, which all exhibit grokking, the Spearman rank correlation between both training/validation accuracy and training mean local dimension is negative.

\subsection{Local Dimensions Detect Exhaustion of Training Capabilities}
\label{sec:local_dimensions_dialogue_state_tracking}

Finding relevant text segments in response to a query, also called \emph{span prediction} or \emph{sequence tagging}, is a highly relevant problem in various natural language processing settings \citep{jm3}.
One such application is retrieval augmented generation (RAG), where, during the information retrieval step, relevant passages from a large corpus need to be selected \citep{fan2024ragllms} before generating an answer. 
Here we study it in the context of dialogue state tracking---a critical task of dialogue modeling.

\paragraph{Setup}

To demonstrate the application of our local dimension estimates in this setting, we compute them for the sequence-tagging-based dialogue state tracking model \emph{TripPy-R} \citep{heck-etal-2022-robust} trained on the {\MultiWOZ}\texttt{2.1} dataset \citep{eric-etal-2020-multiwoz}.
In dialogue state tracking, the task is to predict the user's intent from the natural language input utterances, and keep track of the user's goal described throughout the conversation by updating the dialogue state \citep{young2010hidden}.
The contextualized hidden states for computing our measure are taken from the last layer of the unified encoder component, which has the {\RoBERTa} architecture.
The actual state tracking is performed based on classifier outputs derived from this encoder's output.
The encoder is fine-tuned during the TripPy-R training via the loss signal derived from predicting the current dialogue turn's state update.
We train the models for 20 epochs with Adam, with a linear learning rate schedule up to \(5 \cdot 10^{-5}\) that starts with one warm-up epoch.

The input data for the local dimension estimates is formatted in the same way as it is used during training and inference. 
A single input sequence consists of a dialogue turn, followed by the dialogue history, where the different sequence components are separated via special tokens.
For the local dimension estimates, we sample \(\DatacorpusSubsampleSize=\num{7000}\) sequences from the training, validation, and test split, with a token subsample size \(\TokenVectorSubsampleSize=\num{60000}\), and neighborhood size \(\LocalNeighborhoodSize=\num{128}\). 

%
\begin{figure}[ht]
    %
    \centering
    \begin{subfigure}{0.315\linewidth}
        \centering
        \includegraphics[
            height=3.1cm,
            trim=0 1mm 15mm 0, 
            clip
        ]{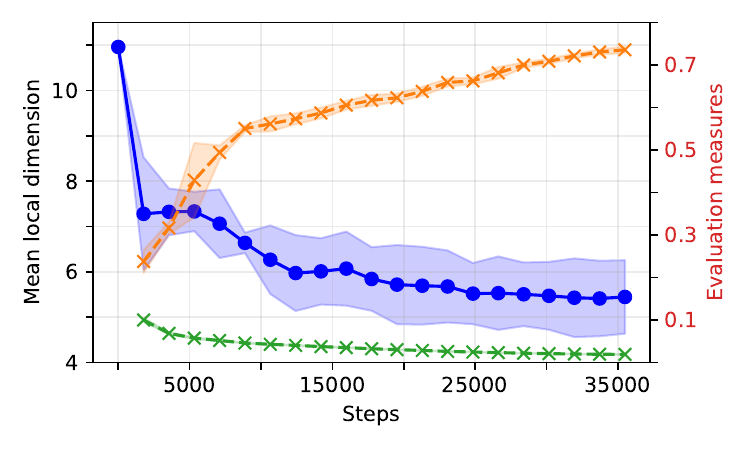}
        \caption{
            {\MultiWOZ} Training split
            \label{fig:trippy_r_multiwoz_20_epochs_linear_learning_rate_schedule_multiwoz_training}
        }
    \end{subfigure}
    \hfill
    \begin{subfigure}{0.315\linewidth}
        \centering
        \includegraphics[
            height=3.1cm,
            trim=15mm 1mm 15mm 0, 
            clip
        ]{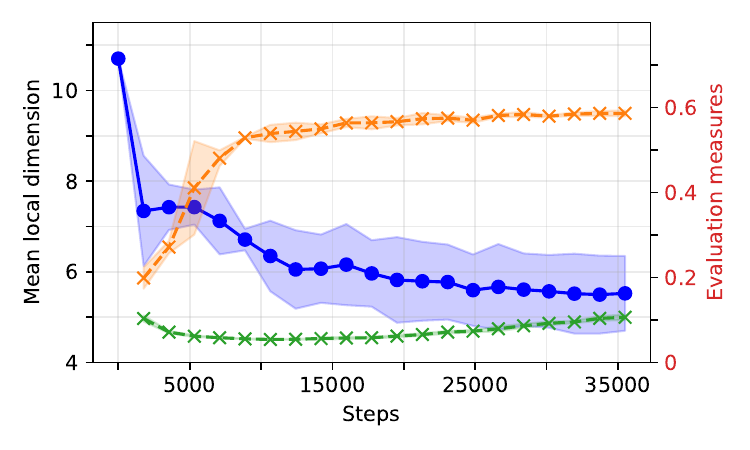}
        \caption{
            {\MultiWOZ} Validation split
            \label{fig:trippy_r_multiwoz_20_epochs_linear_learning_rate_schedule_multiwoz_validation}
        }
    \end{subfigure}
    \hfill
    \begin{subfigure}{0.315\linewidth}
        \centering
        \includegraphics[
            height=3.1cm,
            trim=15mm 1mm 0 0, 
            clip
        ]{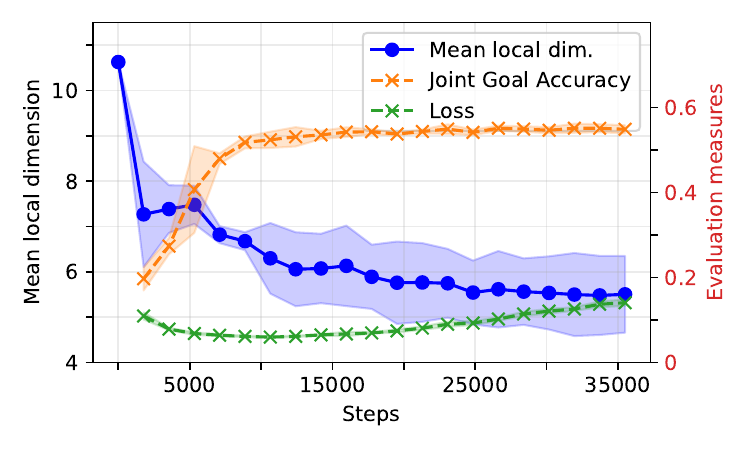}
        \caption{
            {\MultiWOZ} Test split
            \label{fig:trippy_r_multiwoz_20_epochs_linear_learning_rate_schedule_multiwoz_test}
        }
    \end{subfigure}
    \caption{
        Development of TripPy-R performance measures (model loss in green; joint goal accuracy in orange) compared with mean local dimension estimates (blue) evaluated on the training, validation, and test split of the {\MultiWOZ} dataset.
        We show the mean and standard deviation of the measures evaluated at the end of each epoch over six different model training seeds.
        \label{fig:trippy_r_multiwoz_20_epochs_linear_learning_rate_schedule}
    }
\end{figure}

\paragraph{Results}
\Cref{fig:trippy_r_multiwoz_20_epochs_linear_learning_rate_schedule} shows the development of two model performance measures:
One is the differentiable model loss, which is used for backpropagation.
The other is the non-differentiable \emph{joint-goal accuracy} (JGA), which is used to judge the downstream performance of the state tracking model.

Note that the mean local dimension estimates behave similarly when comparing the training, validation, and test splits.
This is in contrast to the loss and JGA:
On the training set, the loss (green) is monotonically decreasing, while the JGA (orange) is increasing.
On the other hand, the loss reaches a minimum on the validation set after \num{7500} batches, long before the JGA converges.
In this case, the differentiable model validation loss would give the wrong impression that the model has finished learning generalizable representations. 
However, from the dimension estimate, we conclude that the model has not converged at that point.
The mean local dimension is still decreasing, and one should continue training.
Notably, we can make this observation on the training data alone and would not need a separate labeled validation set when relying upon our estimates.

Quantitatively, taken over the whole time series progression and step-wise averaged over six seeds, the mean local dimension exhibits a strong negative Spearman rank correlation with the JGA of \(-0.982\)/\(-0.958\)/\(-0.905\) for the training/validation/test split, respectively.
This confirms that progressive compression of the embedding space directly reflects improvements in generalizable task performance.
Correlation between the mean local dimension and the loss is positive \((0.982)\) on the training set, and negative (\(-0.612\)/\(-0.734\)) on the validation and test set, mirroring the fact that the loss does not align consistently with JGA performance over the different splits.

In another direction, we can see that later in the training, after roughly \num{25000} batches, the stabilization of the local dimension estimates coincides with the convergence of the model performance regarding JGA on the validation and test sets.
This effect of stabilization of the local estimates, indicating that the training capabilities have been exhausted, can be seen even more clearly for longer training runs (50 epochs) in \Cref{fig:trippy_r_multiwoz_50_epochs_linear_learning_rate_schedule} in \Cref{appendix:additional_trippy_r}.

\subsection{Local Dimensions Detect Overfitting}
\label{sec:local_dimensions_emotion_recognition}

We now examine the mean local dimension as an indicator of overfitting in a challenging sequence classification setting.

\paragraph{Setup}
We use the {\EmoWOZ} dataset \citep{feng-etal-2022-emowoz}, with the task of classifying dialogue utterances into one of seven emotion classes.
Our experiments employ the ERToD model \citep{feng-etal-2023-chatter} in its baseline configuration, a linear classifier built upon a \texttt{bert-base-uncased} \citep{devlin-etal-2019-bert} model feature extractor.
We intentionally decide on the setup \emph{without} dialogue history, data augmentation, or auxiliary predictions, which are modifications proposed by the authors to solve the problem of classifying rare emotions.
The model is trained for 8 epochs using a linear learning rate schedule with warmup for AdamW and peak learning rate of \(2 \cdot 10^{-5}\).
At the end of each training epoch, local dimension estimates are computed from hidden states extracted at the last BERT-encoder layer.
Thus, this is the latent space that the model has tuned to aid its subsequent classification.
We select representative subsets of \(\DatacorpusSubsampleSize=\num{7000}\) sequences from the training, validation, and test split, with a token subsample size \(\TokenVectorSubsampleSize=\num{60000}\), and neighborhood size \(\LocalNeighborhoodSize=\num{128}\).
Model performance is evaluated using cross-entropy loss and weighted F1 and macro F1 classification scores. 
To better capture effects on minority classes, the majority ``neutral'' class is excluded from the F1-scores.

%
\begin{figure}[t]
    \centering
    \begin{subfigure}{0.325\linewidth}
        \centering
        \includegraphics[height=2.8cm,
            trim=0 0 5mm 0, 
            clip
        ]{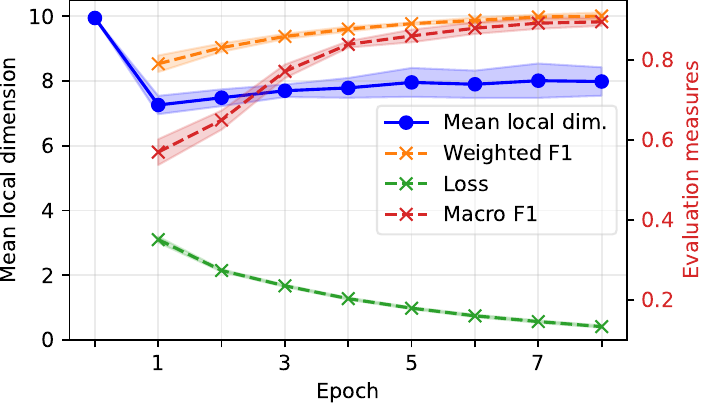}
        \caption{
            {\EmoWOZ} Training split
            \label{fig:emotion_experiments_main_plot_emowoz_training}
        }
    \end{subfigure}
    \hfill
    \begin{subfigure}{0.325\linewidth}
        \centering
        \includegraphics[height=2.8cm,
            trim=5mm 0 5mm 0, 
            clip
        ]{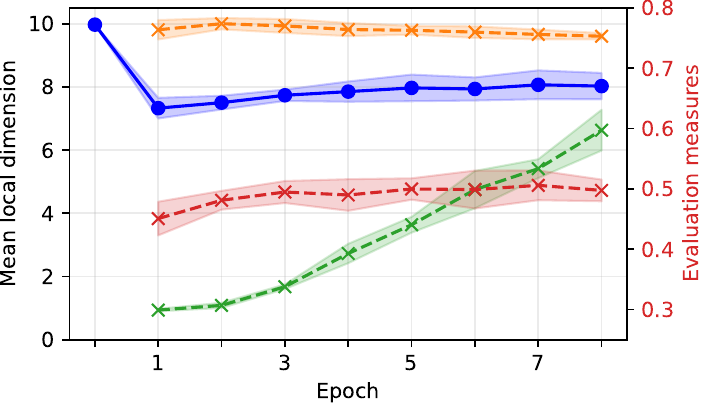}
        \caption{
            {\EmoWOZ} Validation split
            \label{fig:emotion_experiments_main_plot_emowoz_validation}
        }
    \end{subfigure}
    \hfill
    \begin{subfigure}{0.325\linewidth}
        \centering
        \includegraphics[height=2.8cm,
            trim=5mm 0 0 0, 
            clip
        ]{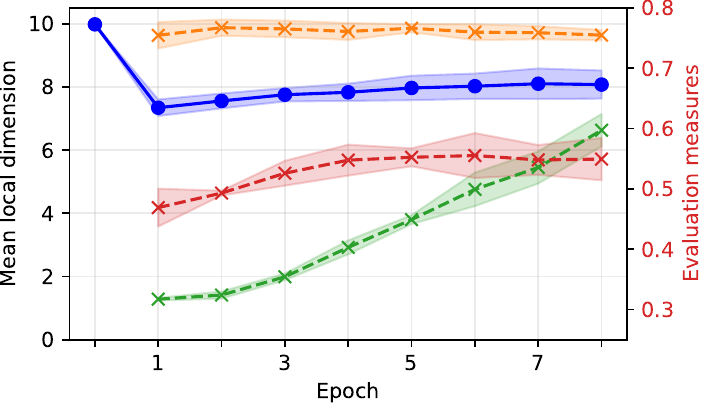}
        \caption{
            {\EmoWOZ} Test split
            \label{fig:emotion_experiments_main_plot_emowoz_test}
        }
    \end{subfigure}
    \caption{
        Development of emotion recognition model performance measures (loss in green; weighted F1 in orange; macro F1 in red) compared with mean local dimension estimates (blue) evaluated on the training, validation, and test split of the {\EmoWOZ} dataset.
        We show the mean and standard deviation of the measures evaluated at the end of each epoch over four different training seeds.
        \label{fig:emotion_experiments_main_plot}
    }
\end{figure}

\paragraph{Results}

\Cref{fig:emotion_experiments_main_plot} depicts the model's performance measures plotted against the mean local estimates.
After the first training epoch, the model's mean local dimension drops notably to \(\approx 7.25\) from the base model's initial dimension of \(\approx 9.94\). 
Over the subsequent eight training epochs, this dimension increases to a value of about 8.
Note that this behavior of the mean local estimates can be observed on all three data splits.
As expected, the training loss monotonically decreases, while both weighted and macro F1-scores increase on the train set.

However, the minimum of the local dimension, followed by an increase, suggests that the model has already found an efficient representation after a single training epoch.
We previously observed a connection between rising dimension and memorization in the grokking experiments in \Cref{sec:local_dimensions_grokking}.
This suspicion is confirmed when considering the performance measures on the validation set in \Cref{fig:emotion_experiments_main_plot_emowoz_validation}.
After the first epoch, the validation loss increases strongly over the following epochs; thus, the model is clearly \emph{overfitting} on the training examples.

In terms of how this influences the classification accuracy, the effect is more nuanced. 
While the macro F1-score increases and plateaus after about three epochs, the weighted F1-score declines from the end of epoch two onward.
This combined behavior implies that the model improves on minority classes at the expense of performance on the majority classes, potentially memorizing rare instances and diminishing its generalization capability.
This interplay between macro and weighted F1 suggests that the model is learning a decision boundary that trades off the performance of majority and minority classes, a phenomenon that may be overlooked when looking at the loss curve alone. 
Practitioners monitoring the shifts of local dimensionality on the training set could, therefore, identify subtler model behavior \emph{without} requiring labeled validation or test data.

Note that in general, the evaluation criteria are task-specific:
In contrast to the observations in \Cref{sec:local_dimensions_dialogue_state_tracking}, where the validation loss is not a reliable stopping criterion when aiming for the target metric JGA, here, the validation loss and weighted F1 are tightly aligned.
Hence, the loss serves as a valid stopping signal here.
The local dimension measures support this point, since they are negatively correlated with the training loss (Spearman rank \(-0.952\)), but positively correlated with the validation (\(0.952\)) and test loss (\(0.976\)). 

\subsection{Computational Resources}
\label{sec:computational_resources}


Fine-tuning of the RoBERTa-base and GPT-2-medium models (in \Cref{appendix:additional_finetuning}) on one of our selected datasets can be performed efficiently on a single NVIDIA V100 or GTX1080TI-12GB GPU within a few hours. 
Additionally, computing the embeddings for a single layer of these models requires only forward passes, which takes approximately 10 minutes on the same hardware.

For the {\twonntext} computations, which are CPU-intensive, the computational requirements depend on the size of the dataset and the dimensionality of the embeddings. 
The computation is feasible in 20 minutes on an E5-2640v4 (Broadwell) 2.40GHz dual-core machine with 32GB of RAM using the \texttt{scikit-dimension} package \citep{bac2021scikitdim} for a typical dataset with tens of thousands of points in high-dimensional space (ambient dimension in the hundreds).
For the grokking experiments, where we evaluate the local dimension frequently, this amounts to a total of 48 CPU core hours per run in the depicted range.
However, precise timing may vary depending on implementation optimizations and the chosen neighborhood sizes.
For a constant and small neighborhood size \(\LocalNeighborhoodSize\) and ambient embedding dimension \(d\), the computation of the neighborhoods for a subsample of \(\TokenVectorSubsampleSize\) using exact search \citep{johnson2019billion} takes \(\mathcal{O}(d N^{2})\).
Then, since we already have the ratios of closest distances available for each neighborhood, the {\twonntext} local dimension estimation is a constant overhead of performing a linear fit on values of a transformed cumulative distribution.

\subsection{Limitations}
\label{sec:limitations}

While versatile, our framework has several limitations.
One drawback is its comparatively high computational complexity; to get a reliable dimension estimate, we need to build a sufficiently large sub-sample of the latent space by performing forward passes on the input data and subsequently computing the neighborhoods and estimates.
Being \emph{independent} of the model training procedure, these computations can be performed in parallel to any potential fine-tuning, thus not slowing down the training process.
Further work could focus on making the local estimate computation more efficient and developing local, stable, and computationally efficient descriptors.

An intrinsic limitation arises from the assumptions of our dimension estimates. 
The {\twonntext} estimate, which we  apply locally, is known to accurately reflect the true local dimension only under strong assumptions on the local point sampling distribution:
The point sample should come from a Poisson point process on a subspace of uniform dimension. 
Note that these assumptions and locally constant density for the hidden states of transformer models have been empirically demonstrated \citep{valeriani2024geometry,viswanathan2025geometrytokensinternalrepresentations}.
While the {\twonntext} method is non-parametric, the precise absolute value of the mean {\twonntext} estimate depends on the hyperparameter choices in selecting the latent space sub-sample and neighborhood size.
This implies that our mean of local dimensions is of a ``relative'' nature, because we cannot directly compare the dimensions' values between different model architectures.
However, this is only of secondary importance for our method, as we are more interested in \emph{changes} of dimension than in absolute values. 
Moreover, our framework is general, permitting the use of other estimation methods.
%
Our working hypothesis is that high-dimensional neighborhoods reflect over-parameterized and poorly localized behavior, whereas very low LID indicates over-compression or memorization.
Between these extremes lies a ``sweet spot'' in LID, where representations retain the essential degrees of freedom required for effective generalization.
Empirically, this intermediate range correlates with peak generalization performance.
While this correlation supports the interpretability of our analysis, establishing a causal link between LID reduction and improved generalization remains an open question for future work.

\section{Conclusion}
\label{sec:conclusion}

We introduce a novel geometric perspective on LLM training dynamics by measuring the local intrinsic dimension of contextual token embeddings. 
While the latent spaces of contextual language models exhibit regional variation in dimensionality, the mean local dimension provides a stable, interpretable summary across dataset splits. 
Across diverse tasks, a sustained drop in mean local dimension reliably suggests improved generalization, offering a practical, unsupervised diagnostic signal that, among other things, enables the detection of grokking.
Such a marked reduction in intrinsic dimension in supervised downstream tasks highlights their strong compressive effect on the latent space.
On the other hand, rising dimensions can point to a tradeoff between generalization and specialization, typically observed during overfitting.
Our approach enables monitoring without supervision by labels, which is particularly valuable in low-resource settings where validation labels are limited or unavailable. 
Thus, our findings highlight the utility of geometric descriptors for monitoring and interpreting LLM behavior beyond traditional label-based metrics.

\paragraph{Impact and Future Work}

To our knowledge, we are the first to investigate local dimensional shifts and their implications in training and fine-tuning contextual language models for different downstream tasks.
The findings of this work enhance the ongoing discussion regarding the interpretability, adaptability, and generalizability of LLMs.
Our work provides a foundation for designing better model architectures and developing interventions that utilize the insight that lower intrinsic dimensions benefit machine learning problems. 
Our measures could inform parameter choices such as LoRA-ranks in the spirit of \citep{eddib2024gelorageometricadaptiveranks}, and are broadly applicable for other (post)-training phases, for example, those which involve RL-tuning.
While our current dimension estimation method is not differentiable (due to nearest-neighbor graph construction and the subsequent {\twonntext} estimator), we see designing a differentiable surrogate or proxy loss to encourage local compression during training as an exciting opportunity for future work. 
The locality of the dimension estimates promises further applications beyond the scope of the current work. 
One immediate avenue for investigation is to see to what extent the dimension can be used to detect which data the model has been trained on. 
Another important direction would be to find connections between the underlying meaning that the tokens carry and their corresponding dimension estimates.

\newpage
\newpage

\begin{ack}
This work was made possible through the support of the European Research Council (ERC) under the Horizon 2020 research and innovation program (Grant No. STG2018 804636), the Ministry of Culture and Science of North Rhine-Westphalia within the Lamarr Fellow Network, and the Swiss State Secretariat for Education, Research, and Innovation (SERI). 
Computational resources were provided by the Centre for Information and Media Technology at Heinrich Heine University Düsseldorf, and Google Cloud.
\end{ack}

\bibliography{references}

\clearpage

\makeatletter
\ifskipchecklist
\fi
\makeatother

\clearpage

\appendix

\startcontents
\printcontents{}{1}{{%
    \vskip10pt\hrule
    \large\textbf{Appendix~(Supplementary Materials)}\vskip3pt\hrule\vskip5pt}
}


\section{Sensitivity Analysis}
\label{sec:sensitivity_analysis}

The following analysis gives insights into the stability of the sampling process in our pipeline of \autoref{alg:compute_local_estimates}.
In particular, we discuss the influence of the following variations on our final mean local estimate:
\begin{itemize}
    \item In \Cref{sec:dependence_on_hyperparameter_choices}, we show that with increasing sample and neighborhood sizes, the {\twonntext} estimates tend to converge in our setting, and already a small portion of samples suffices for reasonable reliability.
    \item In \Cref{sec:noise_analysis} we show that the mean local estimates are stable under applying noise to the embedding vectors.
    \item In \Cref{sec:masked_versus_regular_token_embeddings}, we show that for masked language models, both masked and regular token embeddings lead to similar mean estimates.
\end{itemize}

All embeddings in the following analyses are computed using the regular token embeddings from the last layer of the {\RoBERTa}-base model.

\subsection{Dependence on Hyperparameter Choices}
\label{sec:dependence_on_hyperparameter_choices}

\paragraph{Dataset sequence sub-sampling (dependence on \(\DatacorpusSubsampleSize\))}

Text corpora data splits usually contain different numbers of text sequences, since, for example, the training dataset is larger than the validation or test dataset.
Here, we present experiments to show that our method can be made stable in the size of text corpus sequence sub-sample \(\DatacorpusSubsampleSize \in \NN\), and how it depends on the seed of the sequence sub-sample.

Note that in our sub-sampling, we shuffle the dataset split using a given seed, and then truncate to a beginning segment of the specified size.
Thus, when comparing different data sequence sub-sampling sizes for a fixed seed, the smaller sequence sub-samples are subsets of the larger ones.
In all the experiments in this section, the size of token sub-sample is set to \(\TokenVectorSubsampleSize = \) \num{60000}, and the local neighborhood size parameter \(\LocalNeighborhoodSize = 128\).
Note that for the smallest sequence sub-sample size \(M = \) \num{2000}, not enough non-padding tokens are necessarily available to reach a token sample size of \(N = \) \num{60000}, but with more sequences, we get this common token space size.

\begin{figure}[ht]
    \centering
    \begin{subfigure}{0.5\textwidth}
        \centering
        \includegraphics[width=\linewidth]{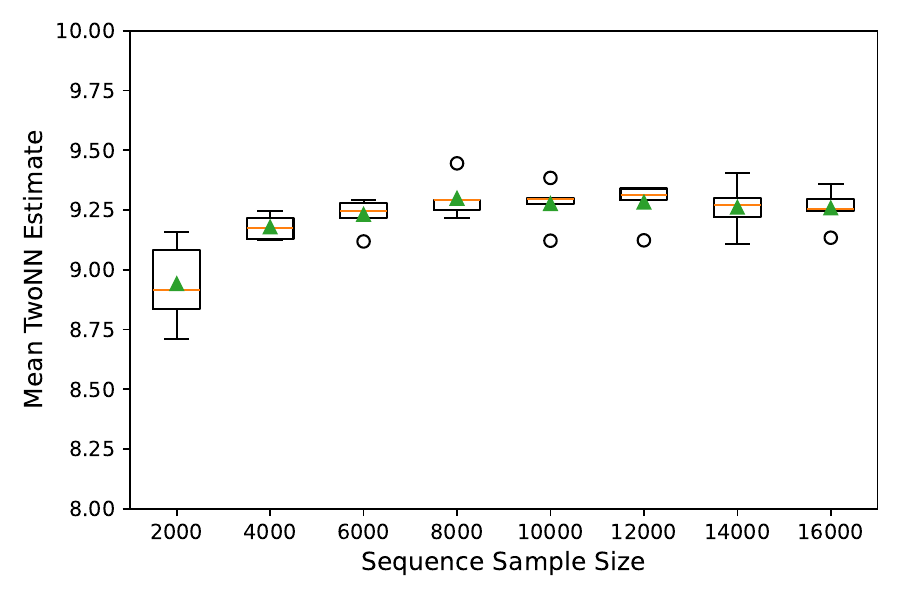}
        \caption{
            {\MultiWOZ} Training split
            \label{fig:sequence_subsampling_multiwoz_training}
        }
    \end{subfigure}
    \hfill
    \begin{subfigure}{0.5\textwidth}
        \centering
        \includegraphics[width=\linewidth]{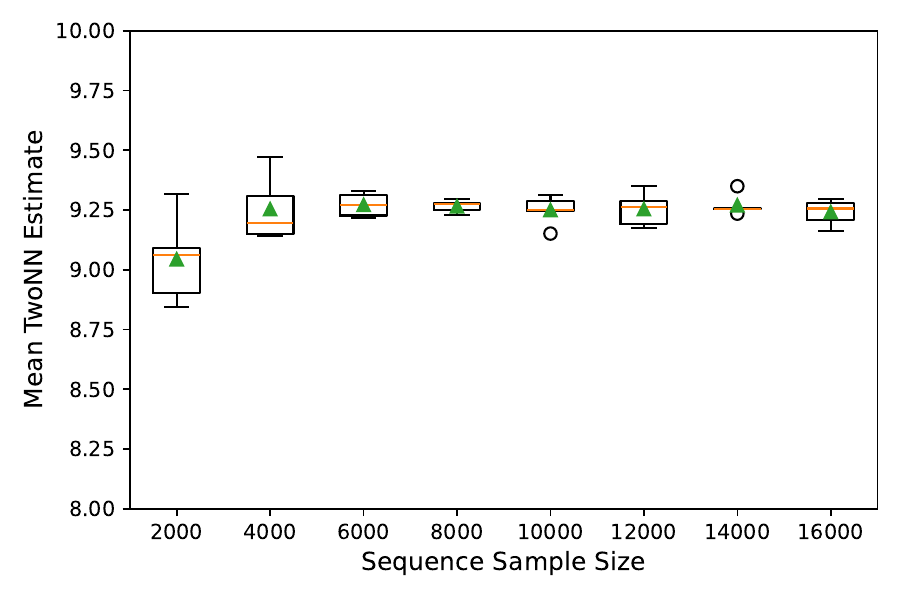}
        \caption{
            {\MultiWOZ} Validation split
            \label{fig:sequence_subsampling_multiwoz_validation}
        }
    \end{subfigure}
    \hfill
    \begin{subfigure}{0.5\textwidth}
        \centering
        \includegraphics[width=\linewidth]{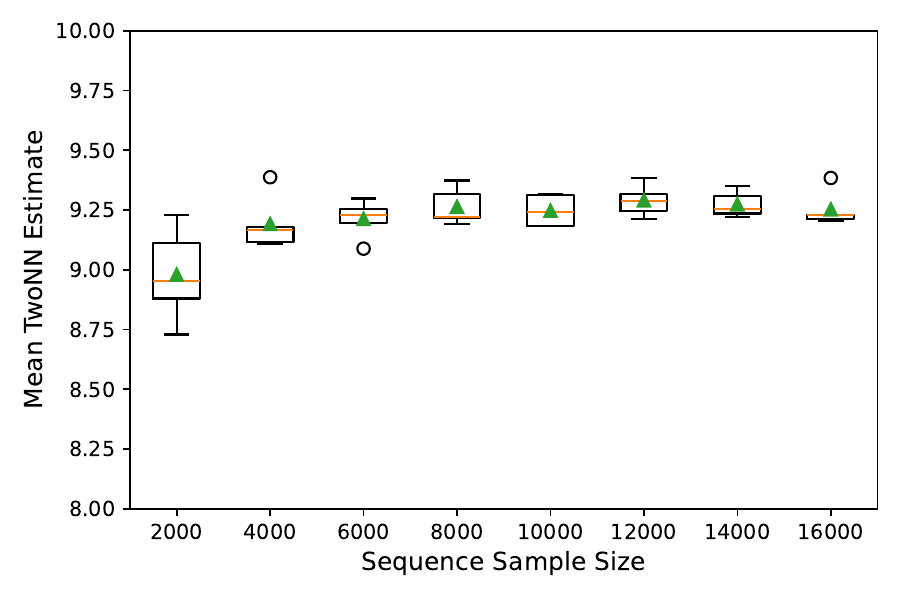}
        \caption{
            {\MultiWOZ} Test split
            \label{fig:sequence_subsampling_multiwoz_test}
        }
    \end{subfigure}
    \caption{
        Boxplots of the mean local {\twonntext} estimates for different sequence sample sizes \(\DatacorpusSubsampleSize\) ranging from \num{2000} to \num{16000}.
        We compute the mean estimates for 5 different sequence sub-sampling seeds for the three splits of the {\MultiWOZ} dataset. 
        Here and in all subsequent boxplots, the average of the mean estimates is depicted as green triangles, their median is the orange line; outliers are shown as circles.
        \label{fig:sequence_subsampling_multiwoz}
    }
\end{figure}

\begin{figure}[ht]
    \centering
    \begin{subfigure}{0.5\textwidth}
        \centering
        \includegraphics[width=\linewidth]{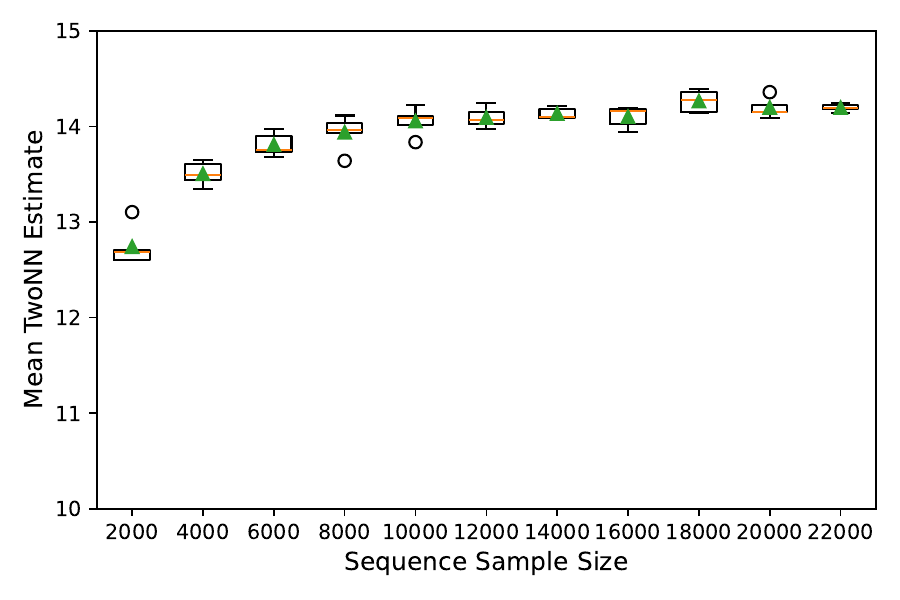}
        \caption{
            {\Reddit} Training split
            \label{fig:sequence_subsampling_reddit_training}
        }
    \end{subfigure}
    \hfill
    \begin{subfigure}{0.5\textwidth}
        \centering
        \includegraphics[width=\linewidth]{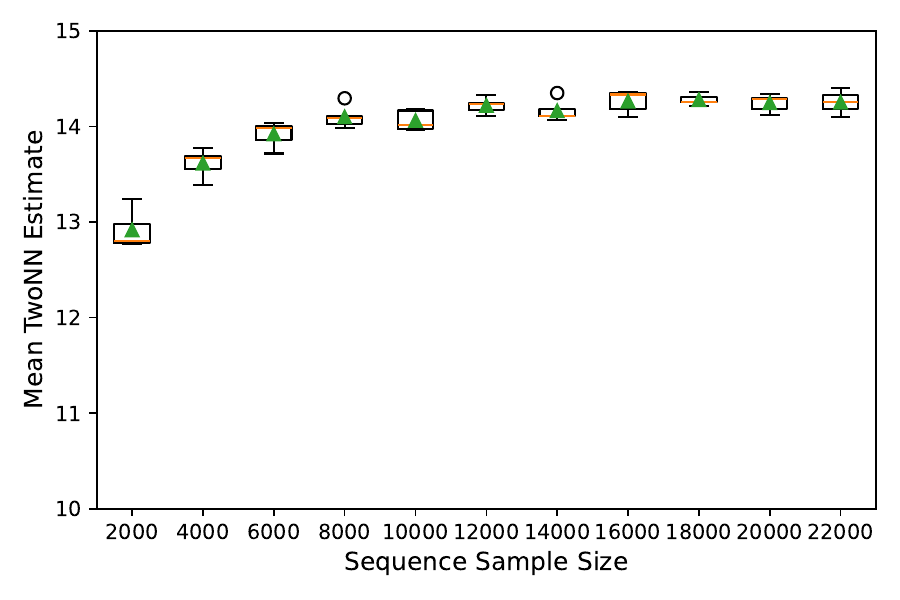}
        \caption{
            {\Reddit} Validation split
            \label{fig:sequence_subsampling_reddit_validation}
        }
    \end{subfigure}
    \hfill
    \begin{subfigure}{0.5\textwidth}
        \centering
        \includegraphics[width=\linewidth]{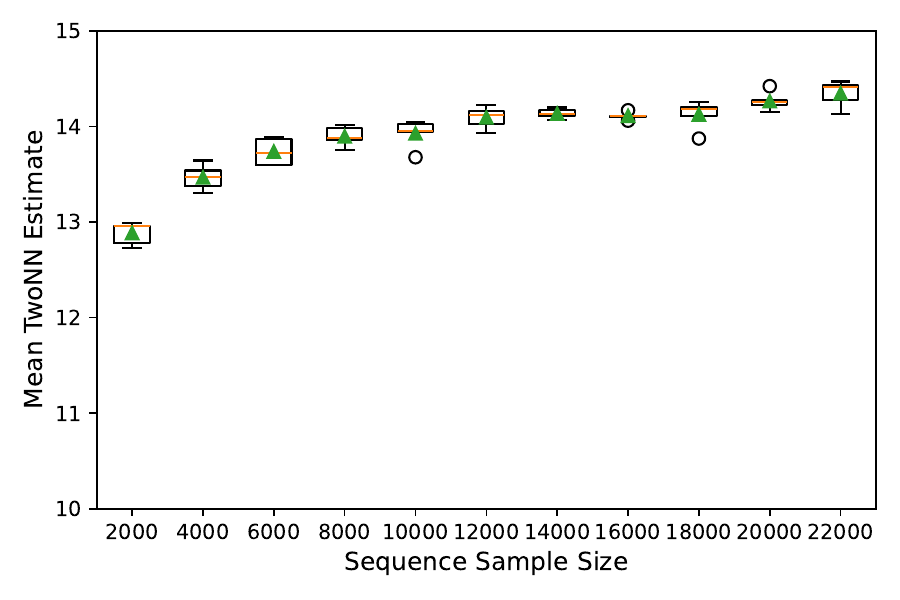}
        \caption{
            {\Reddit} Test split
            \label{fig:sequence_subsampling_reddit_test}
        }
    \end{subfigure}
    \caption{
        Mean local estimates for different sequence sample sizes \(\DatacorpusSubsampleSize\) for splits of the {\Reddit} dataset, shown for 5 sampling seeds.
        \label{fig:sequence_subsampling_reddit}
    }
\end{figure}

We take sequence sub-samples of each given size with 5 different seeds and select a random token sub-sample for each of these, on which the local {\twonntext} estimates are based.
For the results, see \Cref{fig:sequence_subsampling_multiwoz} and \Cref{fig:sequence_subsampling_reddit}.

For a given sequence sample size, the average of the mean local estimates, calculated across different sampling seeds, remains consistent between the various dataset splits of the same dataset. 
This observation indicates that the sampling process is robust to differences in how the data is divided. 
Furthermore, the estimates stabilize as the sequence sample size increases, suggesting convergence of the local mean estimates with larger samples. 
Based on these observations, we set \(\DatacorpusSubsampleSize\) to the values \num{7000} or \num{10000} for the natural language-based tasks in \Cref{sec:local_dimensions_language_modeling_finetuning,sec:local_dimensions_dialogue_state_tracking,sec:local_dimensions_emotion_recognition}, because it represents a practical compromise between computation effort and stability of the value at that scale. 
This value provides sufficient sampling density to capture the underlying data distribution comparably across datasets while maintaining computational efficiency.

\paragraph{Token sub-sampling (dependence on \(\TokenVectorSubsampleSize\))}

Here, we want to investigate the influence of the token sub-sample size on the local {\twonntext} estimate.
We conduct our analysis on the {\MultiWOZ} dataset.
Since here we are interested in the effect of the second sub-sampling process, we want to fix the first sampling step for the sequences:
We take the sub-sample of the first \(\DatacorpusSubsampleSize =\) \num{10000} sequences from the {\MultiWOZ} training, validation, and test split.
From the resulting token embedding spaces, we then sample \num{100000} points (ignoring padding and end-of-sequence (EOS) tokens).
Finally, we compute the local estimates based on the first \(N = \) \num{10000}, \num{20000}, \ldots, \num{90000}, \num{100000} points in the given sub-sample.
Note that we are left with slightly fewer points in the local estimation step for the largest sub-sample size because of our de-duplication step in \autoref{alg:compute_local_estimates}.
This means that in a given token sub-sample, the beginning segment of the first \num{10000} points is the same even when increasing the sample size, and we compare the mean of the local estimates of this beginning segment.

\begin{figure}[ht]
    \centering
    \begin{subfigure}{0.5\textwidth}
        \centering
        \includegraphics[width=1.0\textwidth]{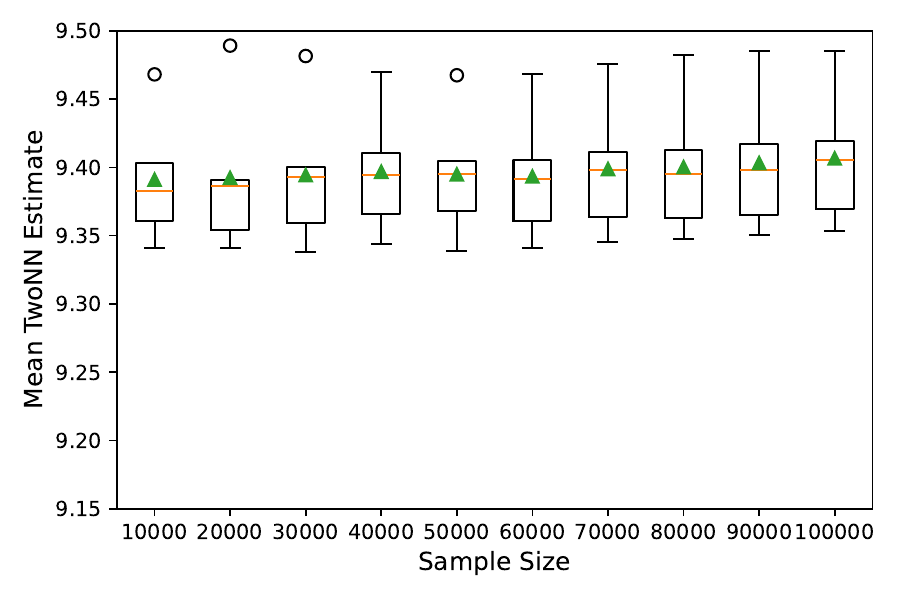}
        \caption{
            {\MultiWOZ} Training split
            \label{fig:token_subsamples_plot_multiwoz_train}
        }
    \end{subfigure}
    \hfill
    \begin{subfigure}{0.5\textwidth}
        \centering
        \includegraphics[width=1.0\textwidth]{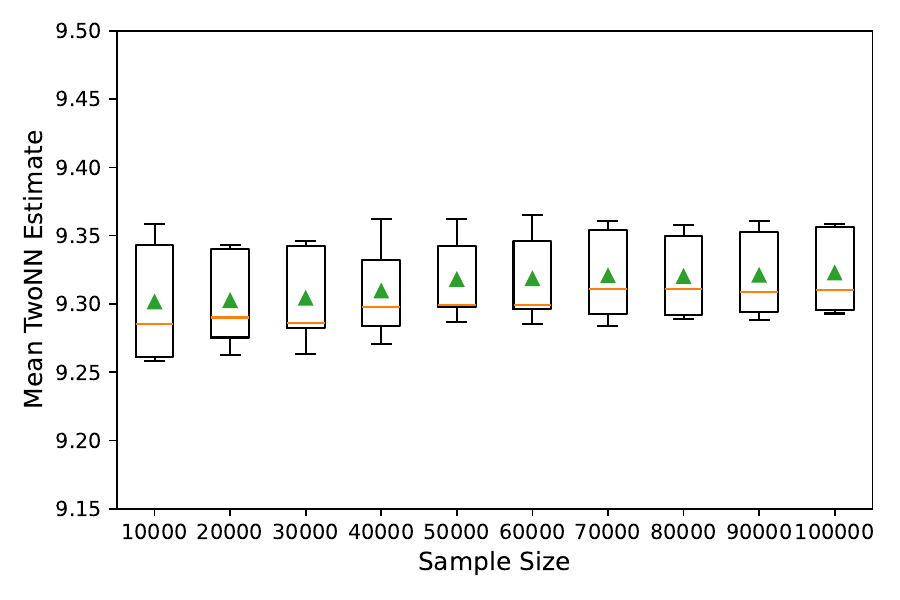}
        \caption{
            {\MultiWOZ} Validation split
            \label{fig:token_subsamples_plot_multiwoz_validation}
        }
    \end{subfigure}
    \hfill
    \begin{subfigure}{0.5\textwidth}
        \centering
        \includegraphics[width=1.0\textwidth]{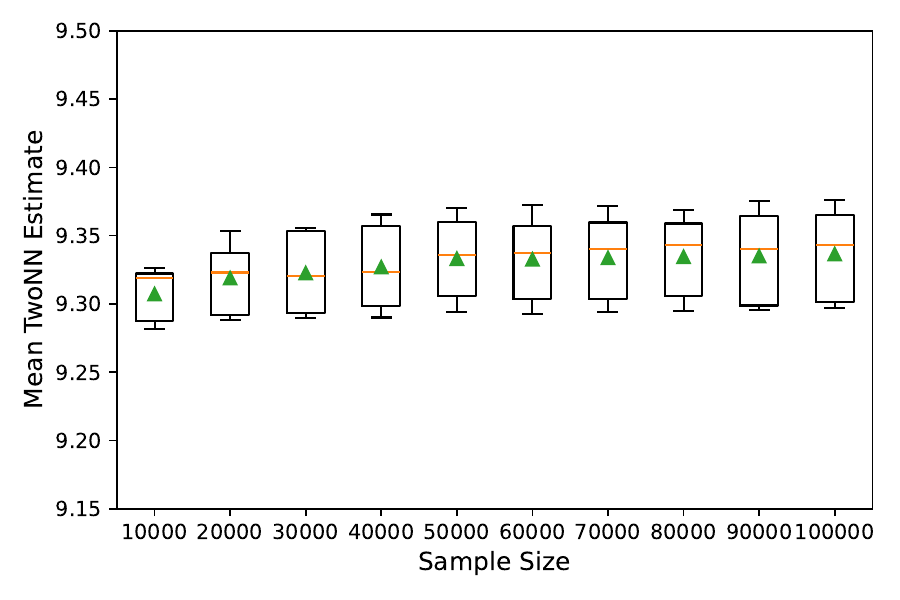}
        \caption{
            {\MultiWOZ} Test split
            \label{fig:token_subsamples_plot_multiwoz_test}
        }
    \end{subfigure}
    \caption{
        Distribution of the mean local {\twonntext} estimates for different cardinalities of token sub-samples.
        We compute the mean estimates for 5 different token sub-sampling seeds and show their distribution for each fixed subsampling cardinality.
        \label{fig:token_subsamples_plot_multiwoz_sensitivity}
    }
\end{figure}

The analysis of the datasets reveals a clear trend in the behavior of the truncated mean values and their associated standard deviations as the sample size increases, see \Cref{fig:token_subsamples_plot_multiwoz_sensitivity}. 
Here, the mean stabilizes quickly, and the standard deviation is comparatively small.
This justifies our choice of \(\TokenVectorSubsampleSize=\num{60000}\) in the natural language dataset settings.

%

\paragraph{Neighborhood size for local estimates (dependence on \(\LocalNeighborhoodSize\))}

\begin{figure}[htb]
    \centering
    \begin{subfigure}{0.4\textwidth}
        \centering
        \includegraphics[width=1.0\textwidth]{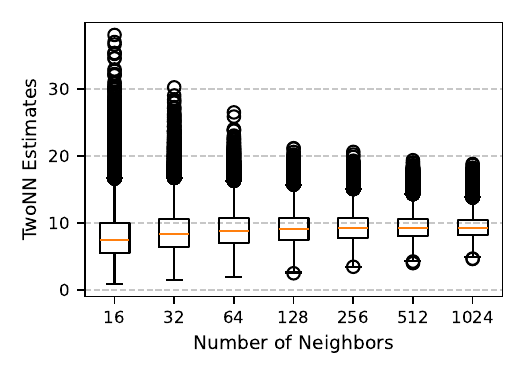}
        \caption{
            {\MultiWOZ} Validation split
            \label{fig:multiwoz_num_neighbors}
        }
    \end{subfigure}
    \hfill
    \begin{subfigure}{0.4\textwidth}
        \centering
        \includegraphics[width=1.0\textwidth]{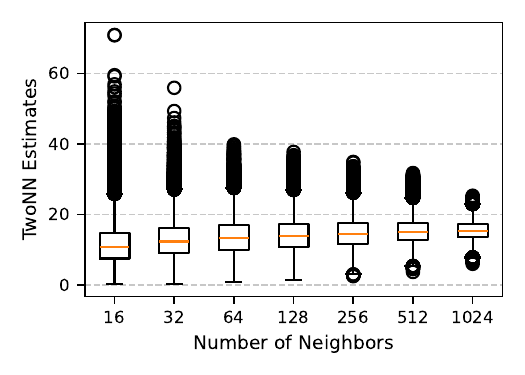}
        \caption{
            {\Reddit} Validation split
            \label{fig:reddit_num_neighbors}
        }
    \end{subfigure}
    \caption{
        Distribution of the local token-wise {\twonntext} estimates over different locality scales \(\LocalNeighborhoodSize\) (number of neighbors).
        The mean and quartiles of the resulting measure varying over the different tokens in a fixed subsample are shown here.
        The more global the estimates, the smaller the standard deviation in the resulting distribution.
        \label{fig:num_neighbor_analysis}
    }
\end{figure}

Having fixed sample sizes \(\DatacorpusSubsampleSize\) and \(\TokenVectorSubsampleSize\), \Cref{fig:num_neighbor_analysis} shows the distribution of {\twonntext} dimension estimates across varying locality scale parameters, represented by the number of neighbors \(\LocalNeighborhoodSize\). 
A key observation is the relative stability of the estimates as the number of neighbors increases, with smaller locality scales (fewer neighbors) yielding more diverse estimate distributions. 
This indicates that local properties of the data significantly influence dimension estimates.

The increasing stability with larger neighborhoods suggests that the embedding space exhibits more homogeneity at broader locality scales. 
However, the variability at smaller scales aligns with the observation in \cite{brown2023verifying} that in many cases, numerical data is not confined to a single manifold of uniform dimensionality but rather consists of multiple manifolds or regions with varying local dimensions.

Our findings provide evidence that token embedding spaces of this nature do not conform to a single, unique dimension. Instead, they exhibit a complex structure, where local estimates diverge, reflecting the diverse geometric and topological properties within the space. This emphasizes the importance of considering locality when evaluating dimension estimates in such embedding spaces.

%

\subsection{Noise Analysis}
\label{sec:noise_analysis}

\begin{figure}[ht]
    \centering
    \begin{subfigure}{0.4\textwidth}
        \centering
        \includegraphics[width=\linewidth]{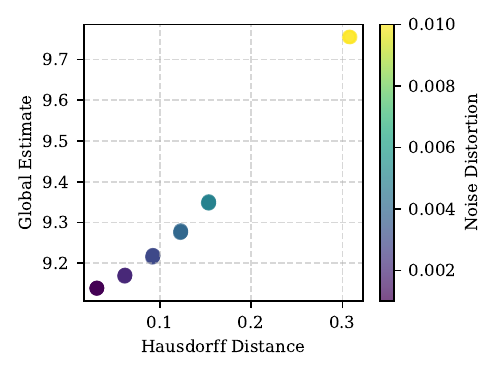}
        \caption{Global {\twonntext} estimate}
        \label{fig:hausdorff_distance_versus_global_twoNN}
    \end{subfigure}
    \hfill
    \begin{subfigure}{0.4\textwidth}
        \centering
        \includegraphics[width=\linewidth]{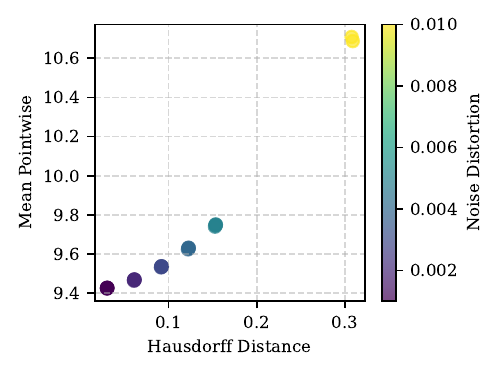}
        \caption{Mean of local {\twonntext} estimates}
        \label{fig:hausdorff_distance_versus_mean_local_twoNN}
    \end{subfigure}
    \hfill
        \begin{subfigure}{0.4\textwidth}
        \centering
        \includegraphics[width=\linewidth]{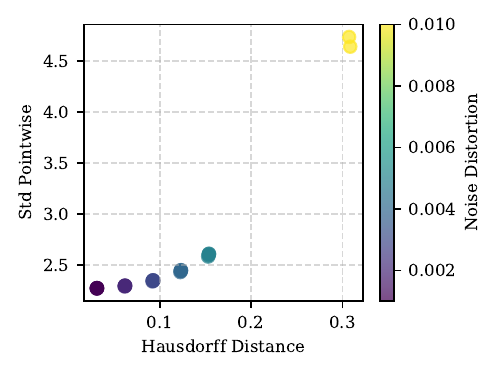}
        \caption{Standard deviation of local {\twonntext} estimates}
        \label{fig:hausdorff_distance_versus_std_local_twoNN}
    \end{subfigure}
    \caption{
        Effect of artificial Gaussian noise on masked token embeddings.
        On the \(x\)-axis, we mark the Hausdorff distance of the noisy embeddings to the clean embeddings,
        with color denoting the distortion \(\sigma\) in the artificial Gaussian noise.
        \label{fig:artificial_gaussian_noise_distances_versus_estimates}
    }
\end{figure}

The following experiments investigate the stability of the {\twonntext} dimension estimates under small perturbations of point clouds in a realistic setting of high-dimensional language model embedding spaces.
The original work in \cite{facco2017estimating} studies the influence of artificial noise on the global {\twonntext} estimator for specific toy datasets (points sampled from a 2-dimensional plane in 20-dimensional space, and points sampled from a 2-dimensional Gaussian distribution wrapped around a Swiss roll).
Here, we confirm the stability under noise for our mean local estimates and in the realistic setting of high-dimensional contextual language model embeddings. 

We analyze the effect of Gaussian noise on the embeddings from the {\MultiWOZ} test set using a fixed set of parameters and varying noise distortions. Specifically, we use \(\DatacorpusSubsampleSize = \num{10000}\) random sequences from the {\MultiWOZ} test set and extract the masked token embeddings from the last layer of the {\RoBERTa} base model, which produces 768-dimensional vectors. 
To compute local estimates, we sub-sample \(\TokenVectorSubsampleSize = \num{60000}\) points from the embeddings and set the neighborhood size to \(\LocalNeighborhoodSize = 128\).

Gaussian noise is applied to all dimensions of the embeddings with varying distortion parameters, \(\sigma \in \{ 0.001 ; 0.002 ; 0.003; 0.004 ; 0.01 \}\), and multiple random seeds to introduce variability. For each noise level, we compute the approximate Hausdorff distance between the point cloud of the \num{60000} clean embedding vectors and the \num{60000} noisy embedding vectors, using the Euclidean distance as the distance metric.

The results are visualized in \Cref{fig:artificial_gaussian_noise_distances_versus_estimates}, where we plot the Hausdorff distance against three measures: the global {\twonntext}-estimate of the noisy point cloud, the mean of the local {\twonntext} estimates, and the standard deviation of the local {\twonntext} estimates. 

From the plots, we observe the following trends. 
The Hausdorff distances between the clean and noisy embeddings and the global and the mean local {\twonntext} estimates increase with larger noise levels, and there is a clear correlation between the two.
Notably, for a distortion parameter of \(\sigma = 0.01\), the difference in the dimension estimates exceeds one. 
As expected, the standard deviation of the local {\twonntext} estimates also increases with higher noise levels, reflecting the increased variability introduced by the noise.
Nevertheless, under small noise levels, the mean local estimate remains relatively stable and has lower variability than the global {\twonntext} estimate.

\subsection{Masked Token Embeddings versus Regular Token Embeddings}
\label{sec:masked_versus_regular_token_embeddings}

\begin{figure}[ht]
    \centering
    \includegraphics[width=0.5\textwidth]{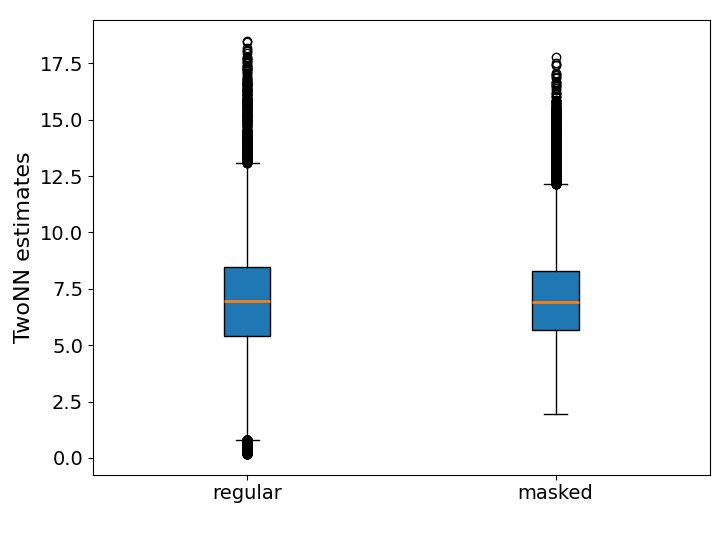}
    \caption{
        {\twonntext} estimates for regular and masked tokens in the {\MultiWOZ} dataset at the last layer. 
        Both token types show similar distributions.
        \label{fig_masked_vs_regular}
    }
\end{figure}

The plots in \Cref{fig_masked_vs_regular} compare the {\twonntext} estimates for regular tokens (left) and masked tokens (right) in the {\MultiWOZ} dataset at the last layer. 
Each box plot represents the distribution of token-wise {\twonntext} estimates.

The estimates for regular tokens have an interquartile range from approximately $5.5$ to $8$, with a median near $7$.
This distribution closely resembles that of masked tokens, which exhibit a similar range and median.
However, the broader spread for regular tokens suggests distinct token-level geometric behaviors influenced by their role in the learned representation space.
In any case, this justifies the computation of the mean local estimates for the regular token embeddings, which are cheaper to obtain:
Every sequence token must be masked individually to retrieve masked embeddings with a corresponding model forward pass. 
In contrast, for the regular embeddings, a single forward pass of the entire sequence automatically produces embeddings for every input token.

\section{Additional Details on Datasets Used}
\label{appendix:additional_details_datasets}

See \Cref{tab:dataset_sizes} for details on the dataset cards and corresponding split sizes.

\begin{table}[htb]
    \caption{
        Dataset sizes for various datasets used in the experiments. 
        Some datasets use pre-determined splits, while others involve random splits.
        \label{tab:dataset_sizes}
    }
    \centering
    \scriptsize
    \begin{tabular}{p{7.5cm} c c c}
        \toprule
        \textbf{Dataset} & \textbf{Training Size} & \textbf{Validation Size} & \textbf{Test Size} \\
        \midrule
        {\MultiWOZ}\texttt{2.1} 
        \linebreak
        \url{https://huggingface.co/datasets/ConvLab/multiwoz21}
        & \num{113556} & \num{14748} & \num{14744} 
        \\
        {\EmoWOZ} 
        \linebreak
        \url{https://paperswithcode.com/dataset/emowoz-1}
        & \num{66474} & \num{8509} & \num{8634} 
        \\
        {\SGD}
        \linebreak
        \url{https://huggingface.co/datasets/google-research-datasets/schema_guided_dstc8}
        & \num{329964} & \num{48726} & \num{84594} 
        \\
        \midrule
        {\ICLR} 
        \linebreak
        submissions (titles + abstracts) 
        & \num{5764} & \num{721} & \num{720} 
        \\
        {{\Reddit} 
        \linebreak 
        \textit{(random 80\%-10\%-10\% split)}
        \linebreak
        {\url{https://huggingface.co/datasets/SocialGrep/one-year-of-tsla-on-reddit}}
        } & \num{174996} & \num{21874} & \num{21874} 
        \\
        {{\Wikipedia} 
        \linebreak 
        \textit{(random 80\%-10\%-10\% split)} 
        \linebreak 
        {\url{https://huggingface.co/datasets/Salesforce/wikitext/viewer/wikitext-103-v1}}} 
        & \num{1441080} & \num{180135} & \num{180135} 
        \\
        \bottomrule
    \end{tabular}
\end{table}

\section{Additional Experimental Results}
\label{appendix:additional_experimental_results}

\subsection{Additional Fine-Tuning Results}
\label{appendix:additional_finetuning}

\paragraph{Fine-tuning on unseen data}

\begin{figure}[t]
    \centering

    \begin{subfigure}[t]{0.9\textwidth}
        \centering
        \includegraphics[width=\linewidth]{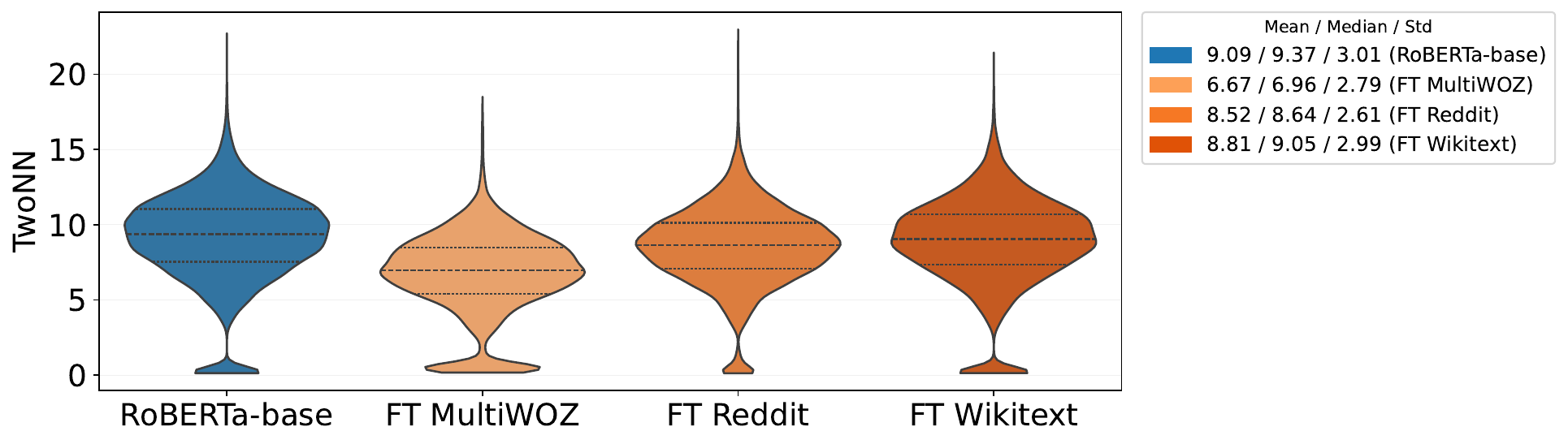}
        \caption{{\twonntext} estimates on {\MultiWOZ} Validation embeddings}
    \end{subfigure}

    \vspace{0.5\baselineskip}
    
    \begin{subfigure}[t]{0.9\textwidth}
        \centering
        \includegraphics[width=\linewidth]{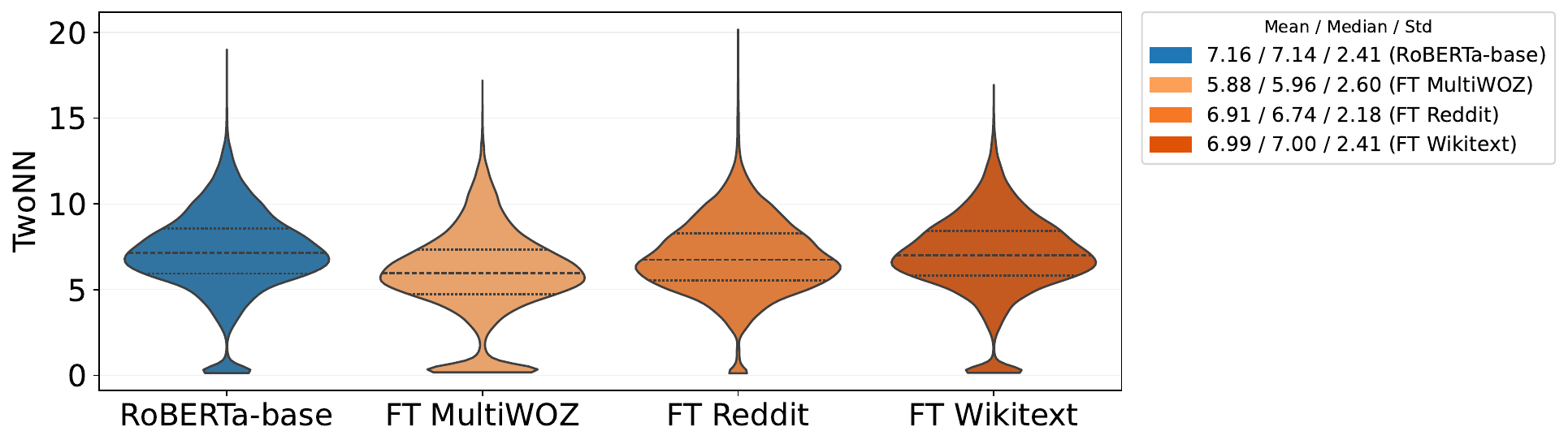}
        \caption{{\twonntext} estimates on {\SGD} Validation embeddings}
    \end{subfigure}

    \vspace{0.5\baselineskip}
    
    \begin{subfigure}[t]{0.9\textwidth}
        \centering
        \includegraphics[width=\linewidth]{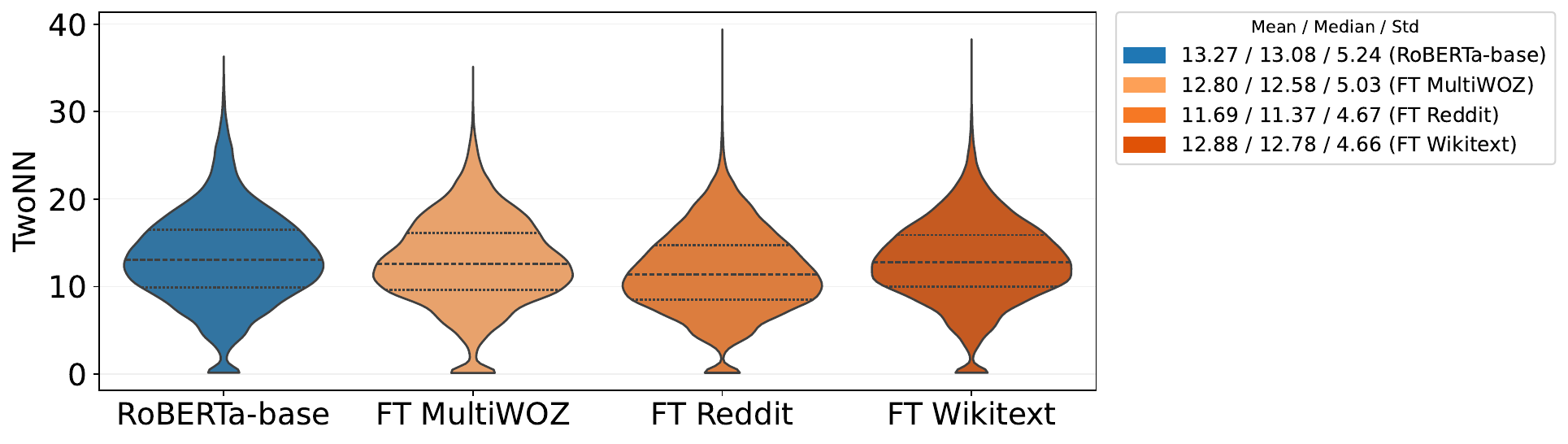}
        \caption{{\twonntext} estimates on {\Reddit} Validation embeddings}
    \end{subfigure}

    \begin{subfigure}[t]{0.9\textwidth}
        \centering
        \includegraphics[width=\linewidth]{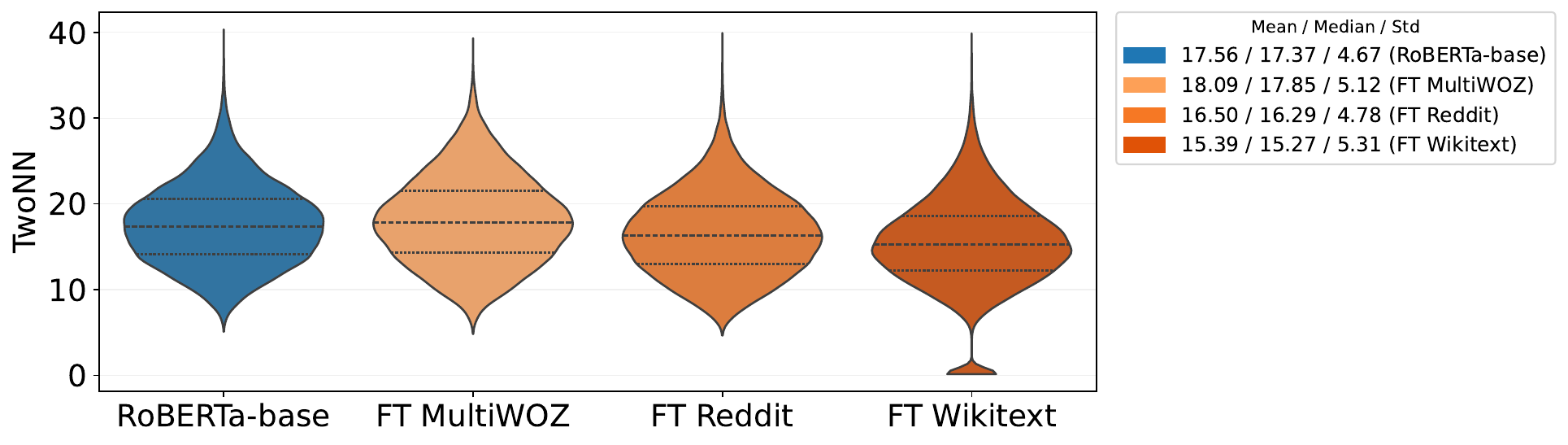}
        \caption{{\twonntext} estimates on {\Wikipedia} Validation embeddings}
    \end{subfigure}

    \caption[Local intrinsic dimension comparison for RoBERTa models]{%
        Distribution of LIDs of the masked language model {\RoBERTa}-base and fine-tunes (\texttt{FT}).
        \label{fig:violin_roberta_multiple_fine_tunings}
    }
\end{figure}

Continuing the discussion in \Cref{sec:local_dimensions_language_modeling_finetuning}, in \Cref{fig:violin_roberta_multiple_fine_tunings}, we provide additional distribution plots of the local estimates for fine-tuned {\RoBERTa}-base models.
Here and in the following, the plots labeled with \texttt{FT} correspond to embedding spaces created from the respective base model fine-tuned on the given dataset ({\MultiWOZ}, {\Reddit}, {\Wikipedia}).

%

\paragraph{Fine-tuning an autoregressive language model}

\begin{figure}[t]
    \centering

    \begin{subfigure}[t]{0.9\textwidth}
        \centering
        \includegraphics[width=\linewidth]{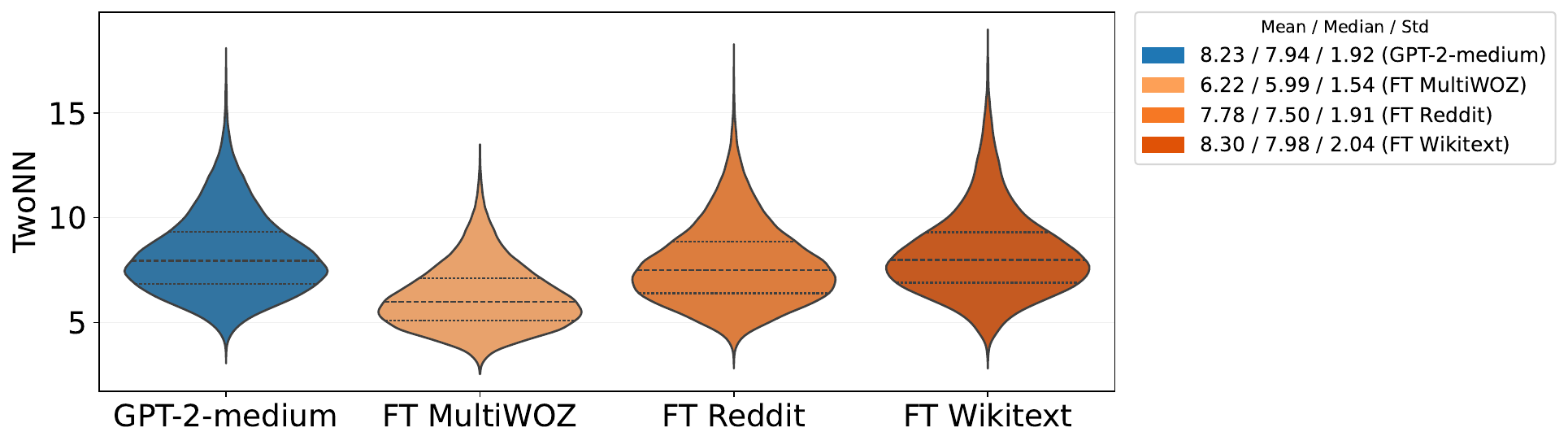}
        \caption{{\twonntext} estimates on {\MultiWOZ} Validation embeddings}
    \end{subfigure}

    \vspace{0.5\baselineskip}
    
    \begin{subfigure}[t]{0.9\textwidth}
        \centering
        \includegraphics[width=\linewidth]{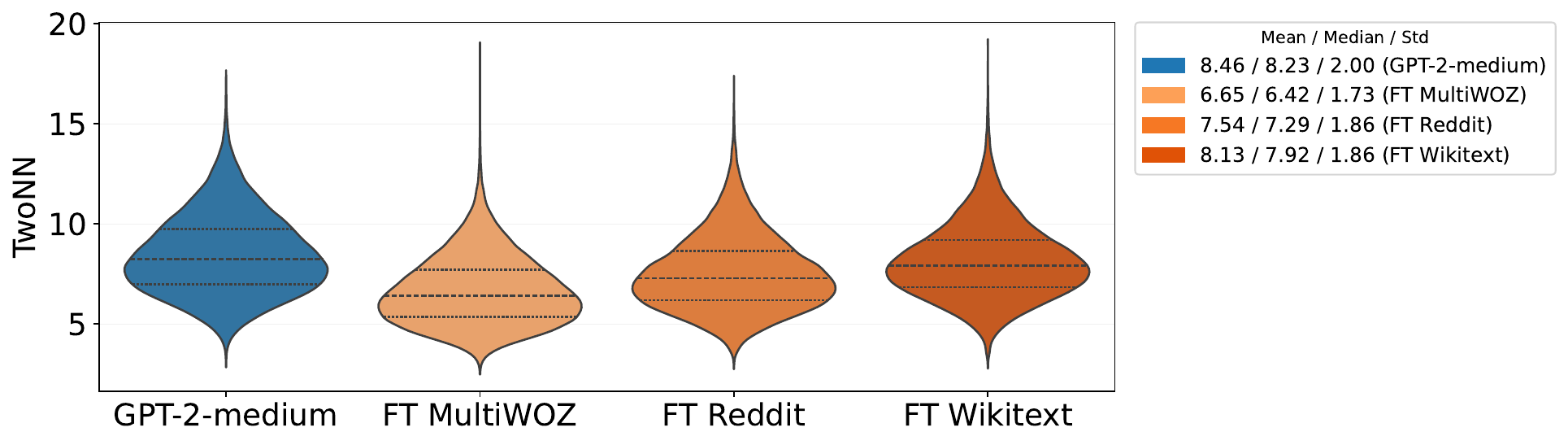}
        \caption{{\twonntext} estimates on {\SGD} Validation embeddings}
    \end{subfigure}

    \vspace{0.5\baselineskip}
    
    \begin{subfigure}[t]{0.9\textwidth}
        \centering
        \includegraphics[width=\linewidth]{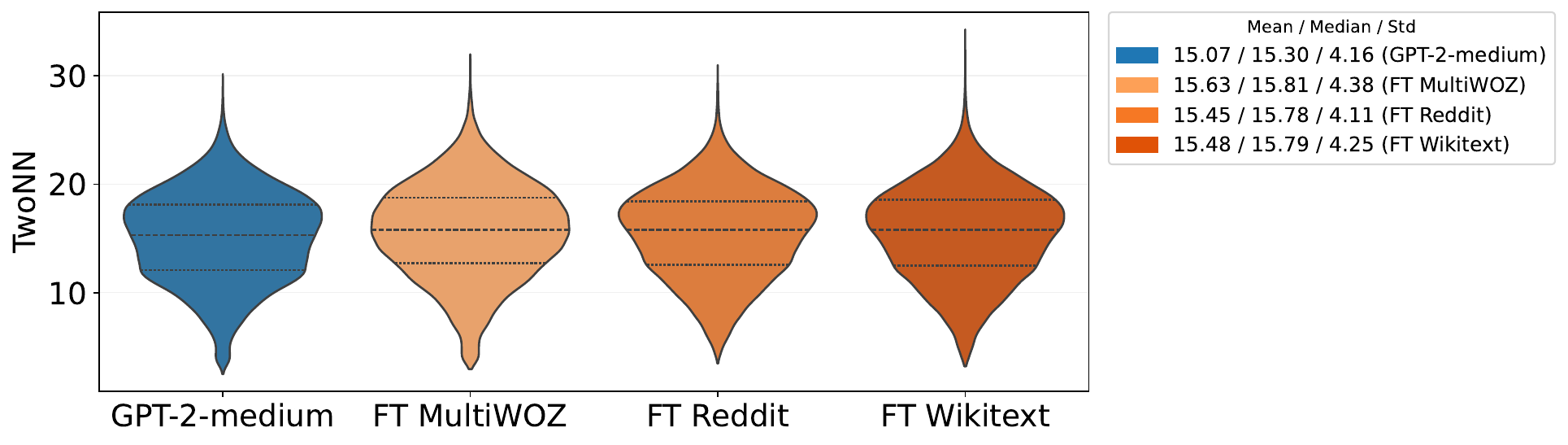}
        \caption{{\twonntext} estimates on {\ICLR} Validation embeddings}
    \end{subfigure}
    
    \begin{subfigure}[t]{0.9\textwidth}
        \centering
        \includegraphics[width=\linewidth]{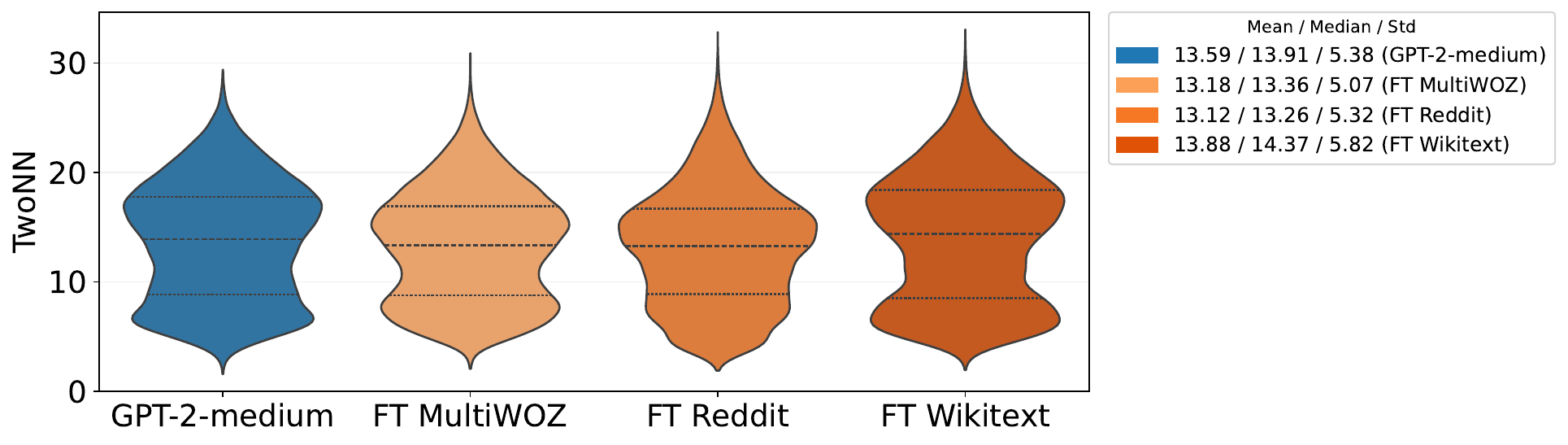}
        \caption{{\twonntext} estimates on {\Reddit} Validation embeddings}
    \end{subfigure}

    \begin{subfigure}[t]{0.9\textwidth}
        \centering
        \includegraphics[width=\linewidth]{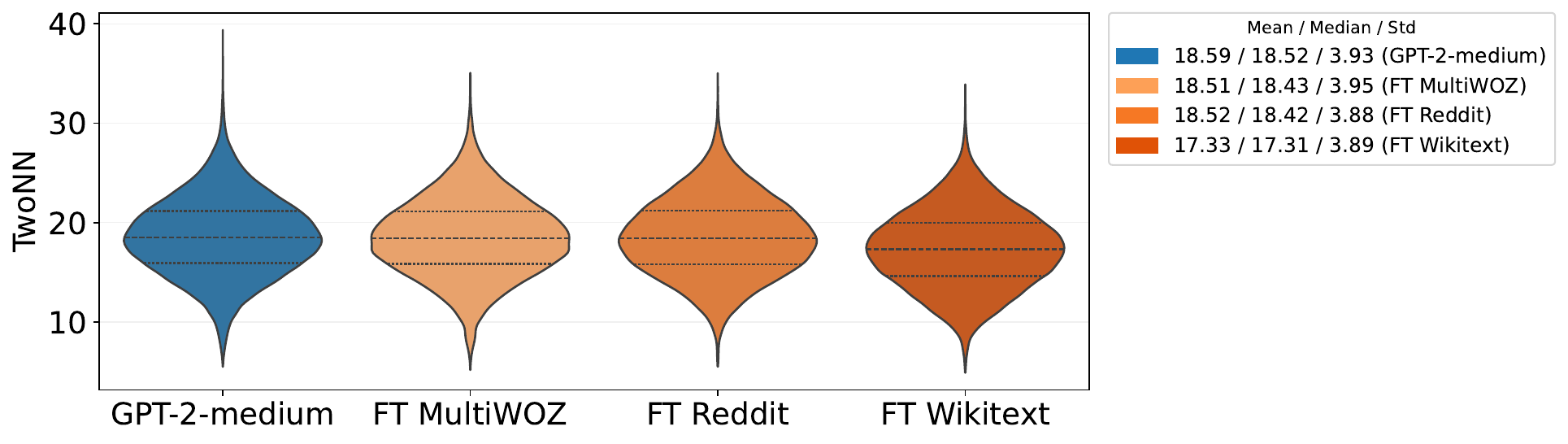}
        \caption{{\twonntext} estimates on {\Wikipedia} Validation embeddings}
    \end{subfigure}

    \caption[Local intrinsic dimension comparison for GPT-2 models]{%
        Distribution of LIDs of the autoregressive {\GPTMedium} model and fine-tunes (\texttt{FT}).
        \label{fig:violin_gpt2_multiple_fine_tunings}
    }
\end{figure}

Moreover, in \Cref{fig:violin_gpt2_multiple_fine_tunings} we show the distribution of local estimates for the auto-regressive {\GPTMedium} model, and their shift under fine-tuning on a causal language modeling task.
The {\GPTMedium} fine-tuning here is performed with standard parameters on the first \num{10000} sequences of the given dataset's train portion, and we show the local estimates at the checkpoint after \num{1200} batches. 

Interesting cases to note are the {\RoBERTa} and {\GPTMedium} models fine-tuned on the {\Reddit} data: 
Since the text from the dataset was created in the year 2022, long after the training of {\RoBERTa} and {\GPTMedium} finished, this combination illustrates the behavior of a model fine-tuned on data that it has not seen during pre-training.
The general observations from \Cref{sec:local_dimensions_language_modeling_finetuning} hold for the autoregressive models as well:
The LIDs of embeddings originating from the fine‑tuning distribution decrease markedly between models, whereas the LIDs for the out‑of‑distribution corpora remain largely unaffected or have increased.

%

\paragraph{Local estimates for larger models}

\begin{figure}[t]
    \centering

    \begin{subfigure}[t]{0.9\textwidth}
        \centering
        \includegraphics[width=\linewidth]{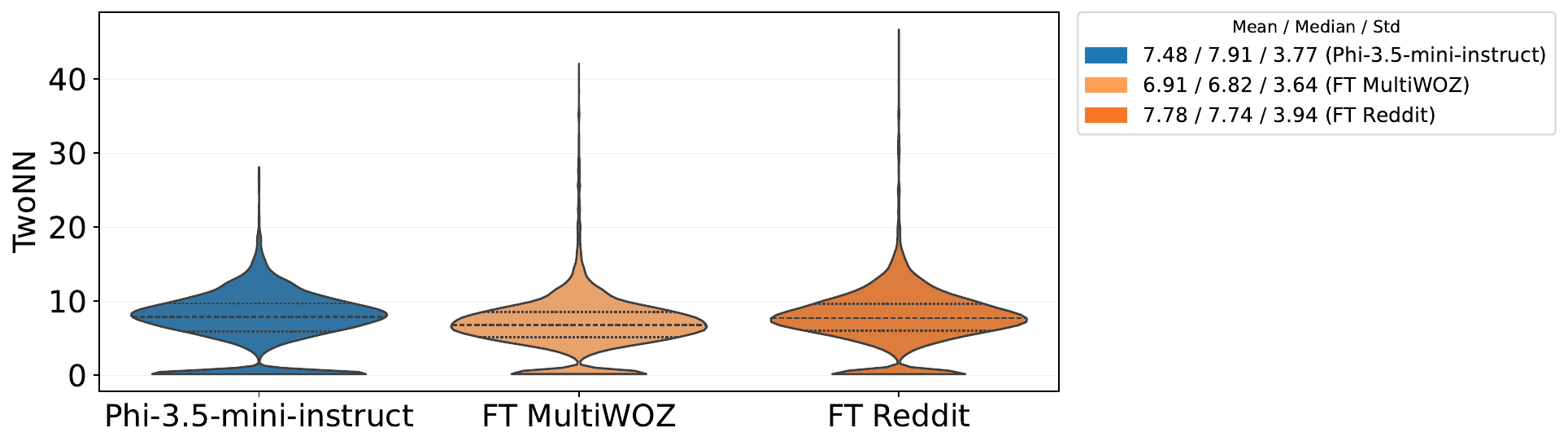}
        \caption{{\twonntext} estimates on {\MultiWOZ} Validation embeddings}
    \end{subfigure}

    \vspace{0.5\baselineskip}
    
    \begin{subfigure}[t]{0.9\textwidth}
        \centering
        \includegraphics[width=\linewidth]{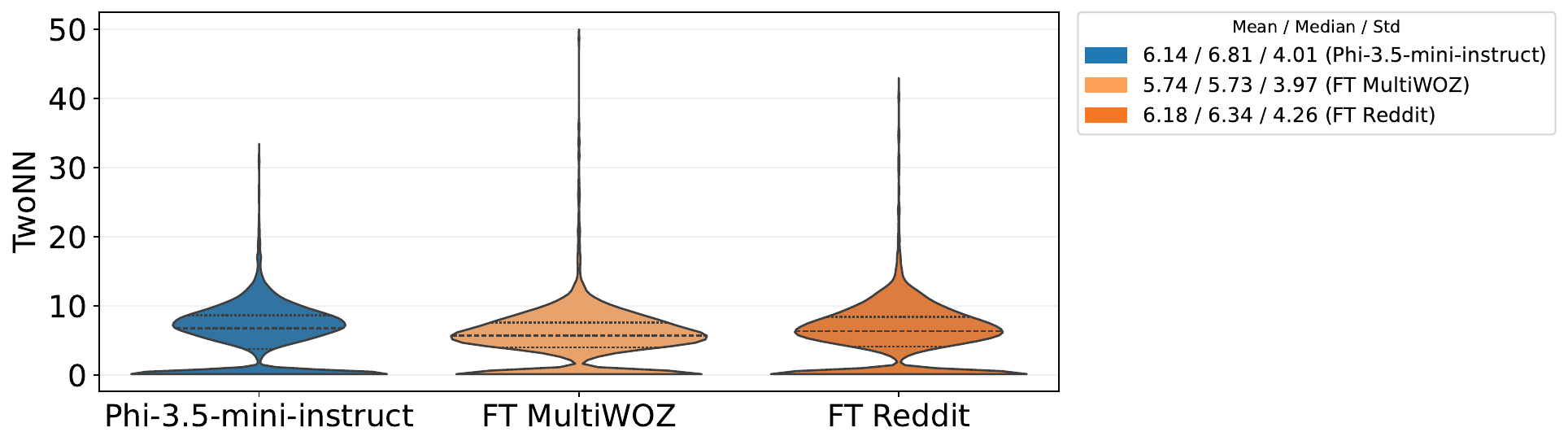}
        \caption{{\twonntext} estimates on {\SGD} Validation embeddings}
    \end{subfigure}

    \vspace{0.5\baselineskip}
    
    \begin{subfigure}[t]{0.9\textwidth}
        \centering
        \includegraphics[width=\linewidth]{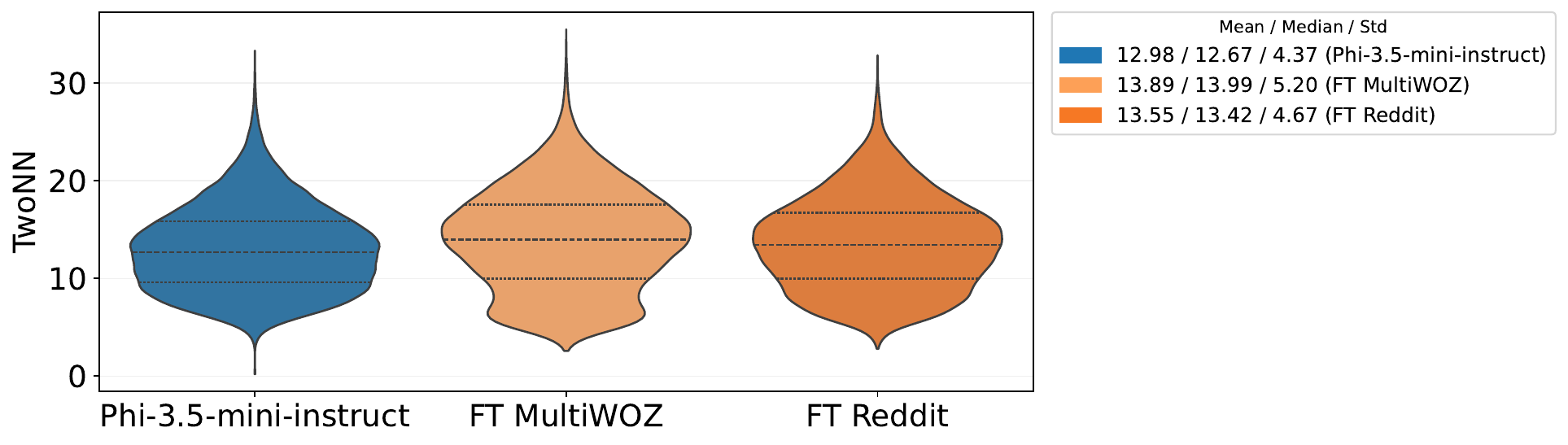}
        \caption{{\twonntext} estimates on {\ICLR} Validation embeddings}
    \end{subfigure}
    
    \begin{subfigure}[t]{0.9\textwidth}
        \centering
        \includegraphics[width=\linewidth]{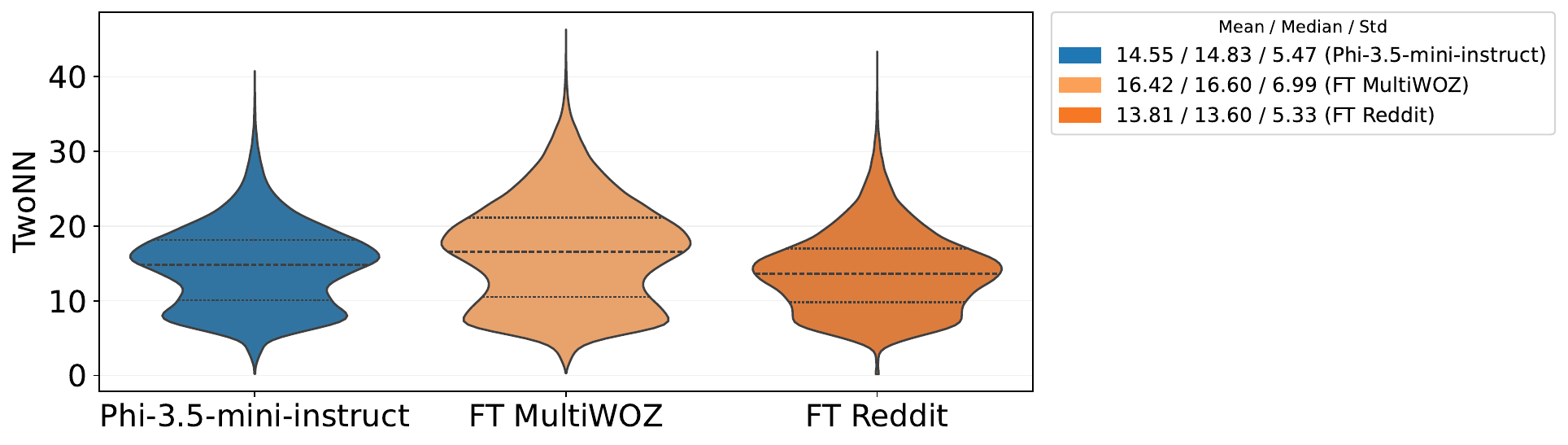}
        \caption{{\twonntext} estimates on {\Reddit} Validation embeddings}
    \end{subfigure}

    \begin{subfigure}[t]{0.9\textwidth}
        \centering
        \includegraphics[width=\linewidth]{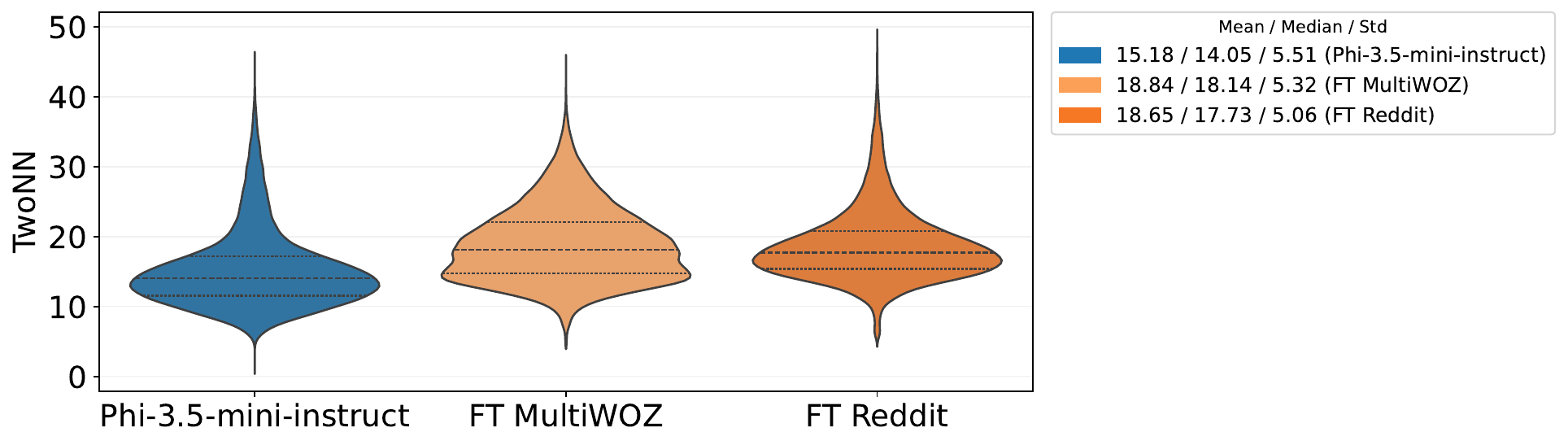}
        \caption{{\twonntext} estimates on {\Wikipedia} Validation embeddings}
    \end{subfigure}

    \caption[Local intrinsic dimension comparison for {\Phiminiinstruct} models]{%
        Distribution of LIDs of the autoregressive {\Phiminiinstruct} model and fine-tunes (\texttt{FT}).
        \label{fig:violin_phi35miniinstruct_multiple_fine_tunings}
    }
\end{figure}

\begin{figure}[t]
    \centering

    \begin{subfigure}[t]{0.9\textwidth}
        \centering
        \includegraphics[width=\linewidth]{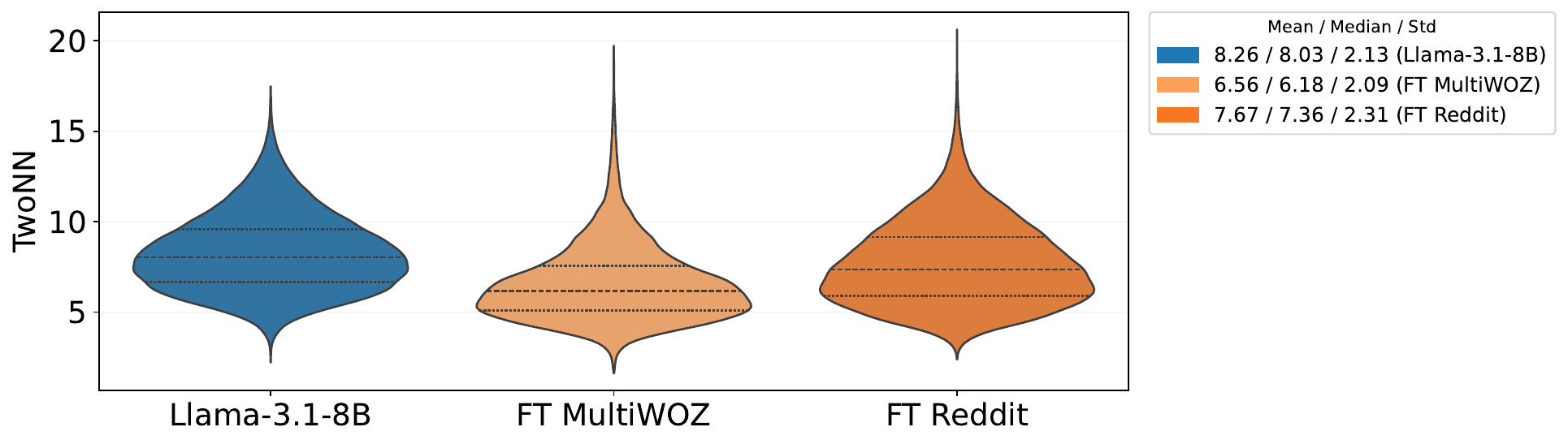}
        \caption{{\twonntext} estimates on {\MultiWOZ} Validation embeddings}
    \end{subfigure}

    \vspace{0.5\baselineskip}
    
    \begin{subfigure}[t]{0.9\textwidth}
        \centering
        \includegraphics[width=\linewidth]{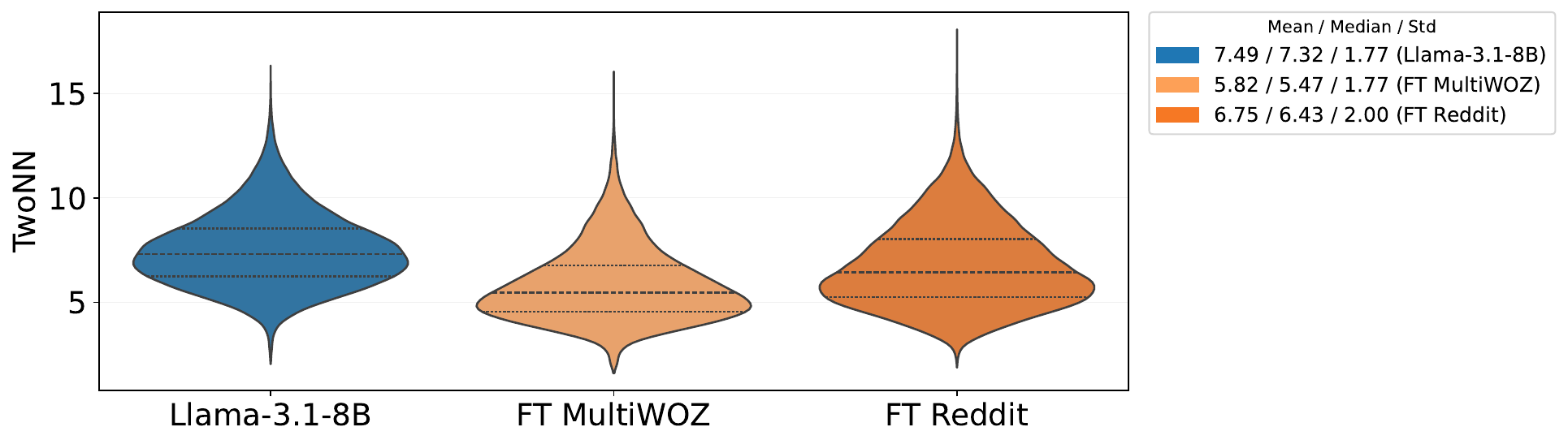}
        \caption{{\twonntext} estimates on {\SGD} Validation embeddings}
    \end{subfigure}

    \vspace{0.5\baselineskip}
    
    \begin{subfigure}[t]{0.9\textwidth}
        \centering
        \includegraphics[width=\linewidth]{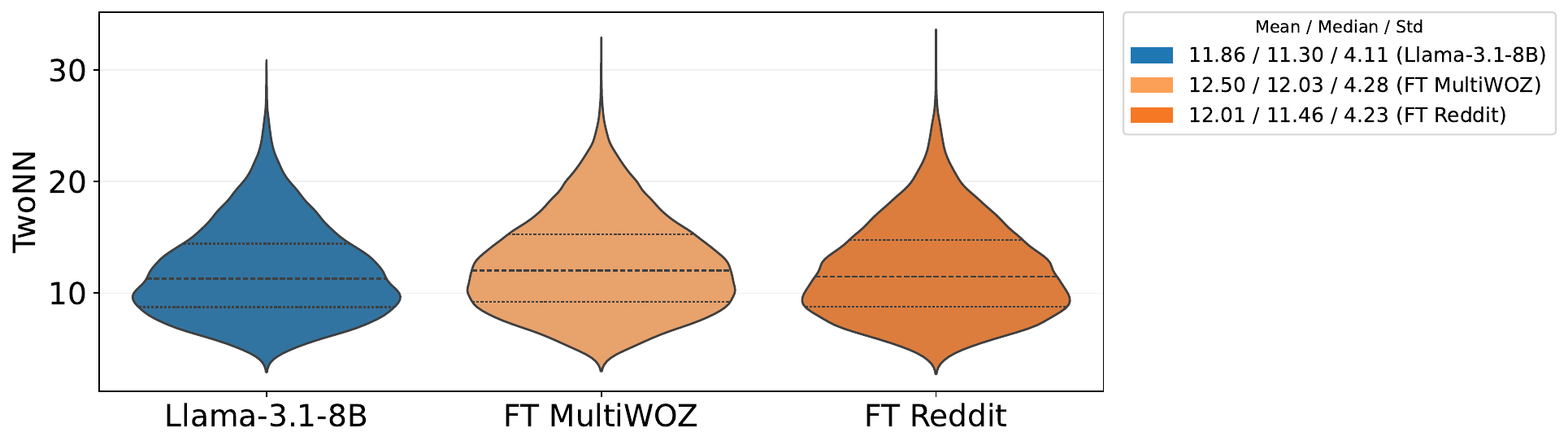}
        \caption{{\twonntext} estimates on {\ICLR} Validation embeddings}
    \end{subfigure}
    
    \begin{subfigure}[t]{0.9\textwidth}
        \centering
        \includegraphics[width=\linewidth]{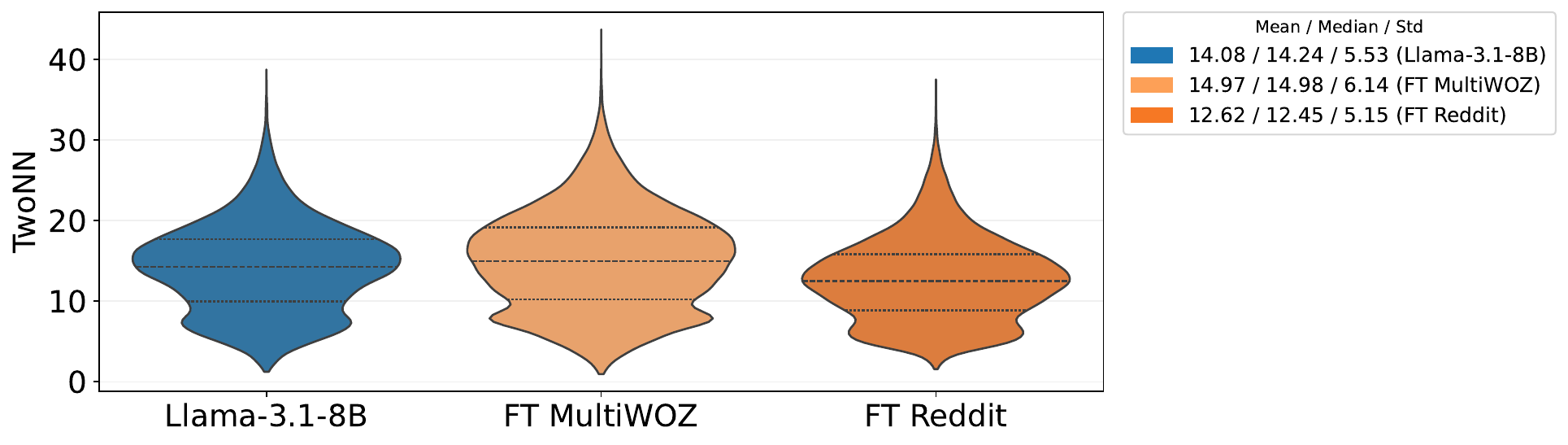}
        \caption{{\twonntext} estimates on {\Reddit} Validation embeddings}
    \end{subfigure}

    \begin{subfigure}[t]{0.9\textwidth}
        \centering
        \includegraphics[width=\linewidth]{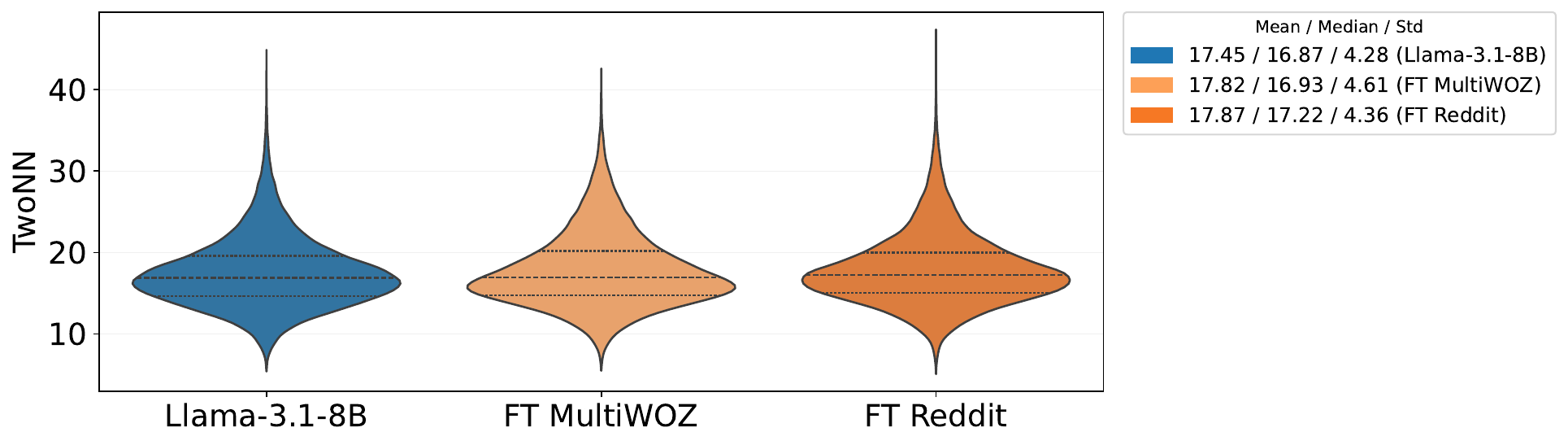}
        \caption{{\twonntext} estimates on {\Wikipedia} Validation embeddings}
    \end{subfigure}

    \caption[Local intrinsic dimension comparison for {\LlamaThreeEightB} models]{%
        Distribution of LIDs of the autoregressive {\LlamaThreeEightB} model and fine-tunes (\texttt{FT}).
        \label{fig:violin_Llama-3.1-8B_multiple_fine_tunings}
    }
\end{figure}

The geometric perspective and methodology we propose are broadly applicable across model architectures.
Local estimates computations for the {\Phiminiinstruct} model in \Cref{fig:violin_phi35miniinstruct_multiple_fine_tunings} and for {\LlamaThreeEightB} in \Cref{fig:violin_Llama-3.1-8B_multiple_fine_tunings} demonstrate that our sub-sampling procedure yields meaningful estimates even for latent spaces whose ambient dimension is in the thousands, a typical setting for modern LLM architectures.

{\Phiminiinstruct} is a decoder model with 3.82B parameters and a hidden dimension of \num{3072} \citep{abdin2024phi3technicalreporthighly}, while {\LlamaThreeEightB} is even larger with 8.03B parameters and a hidden dimension of \num{4096} \citep{llama3modelcard}.
We fine-tune via LoRA with rank constraint \(r=16\) on a random subsample of \num{10000} sequences from the {\MultiWOZ} and {\Reddit} training splits, with other hyperparameters as described in \Cref{sec:local_dimensions_language_modeling_finetuning}.
The violin plots in \Cref{fig:violin_phi35miniinstruct_multiple_fine_tunings,fig:violin_Llama-3.1-8B_multiple_fine_tunings} show the distribution of the LIDs of the validation split for the model checkpoint after 800 batches.

%

%

\subsection{Layer-Wise Computation of Local Estimates}
\label{sec:layerwise_computation}

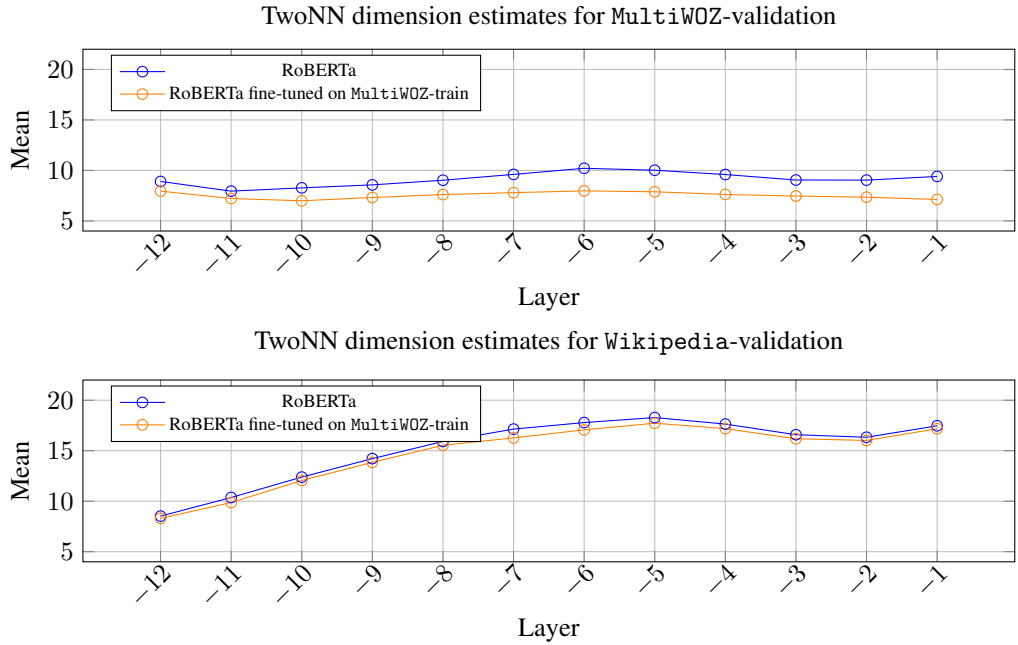
\begin{figure}
    \begin{minipage}{\linewidth}
    \centering
    \begin{tikzpicture}
                \begin{axis}[
                    width=\textwidth,
                    height=4cm,
                    xlabel={Layer},
                    ylabel={Mean},
                    ylabel near ticks,
                    yticklabel style={
                        /pgf/number format/fixed,
                        /pgf/number format/precision=3
                    },
                    ymin=4,
                    ymax=22,
                    grid=both,
                    title={{\twonntext} dimension estimates for \MultiWOZ-validation},
                    legend pos=north west,
                    legend style={nodes={scale=0.7, transform shape}},
                    xtick={1,...,12},
                    xticklabels={
                        \(-12\),
                        \(-11\),
                        \(-10\),
                        \(-9\),
                        \(-8\),
                        \(-7\),
                        \(-6\),
                        \(-5\),
                        \(-4\),
                        \(-3\),
                        \(-2\),
                        \(-1\)
                    },
                    xticklabel style={rotate=45, anchor=north east, inner sep=0mm},
                    ]
                    
                    \addplot+[mark=o, color=blue] table[x index=0, y index=1] {
                        1   8.90931620
                        2   7.95003642
                        3   8.27387782
                        4   8.57033502
                        5   9.03060593
                        6   9.60632770
                        7   10.21180162
                        8   10.02029210
                        9   9.59474338
                        10  9.05242889
                        11  9.04095176
                        12  9.40390724
                    };
                    \addlegendentry{RoBERTa}
                    
                    \addplot+[mark=o, color=orange] table[x index=0, y index=1] {
                        1   7.94831485
                        2   7.21788281
                        3   6.99600488
                        4   7.31947342
                        5   7.60732341
                        6   7.80141745
                        7   7.97543386
                        8   7.88576375
                        9   7.62098645
                        10  7.46393601
                        11  7.35053472
                        12  7.12409058
                    };
                    \addlegendentry{RoBERTa fine-tuned on \MultiWOZ-train}
                    
                \end{axis}
            \end{tikzpicture}
    \par\vfill
    \begin{tikzpicture}
                        \begin{axis}[
                width=\textwidth,
                height=4cm,
                xlabel={Layer},
                ylabel={Mean},
                ylabel near ticks,
                yticklabel style={
                    /pgf/number format/fixed,
                    /pgf/number format/precision=3
                },
                ymin=4,
                ymax=22,
                grid=both,
                title={{\twonntext} dimension estimates for {\Wikipedia}-validation},
                legend pos=north west,
                legend style={nodes={scale=0.7, transform shape}},
                xtick={1,...,12},
                xticklabels={
                        \(-12\),
                        \(-11\),
                        \(-10\),
                        \(-9\),
                        \(-8\),
                        \(-7\),
                        \(-6\),
                        \(-5\),
                        \(-4\),
                        \(-3\),
                        \(-2\),
                        \(-1\)
                },
                xticklabel style={rotate=45, anchor=north east, inner sep=0mm},
                ]
                
                \addplot+[mark=o, color=blue] table[x index=0, y index=1] {
                    1   8.52585749
                    2   10.36608939
                    3   12.38811285
                    4   14.21721963
                    5   15.91846566
                    6   17.14264444
                    7   17.78668704
                    8   18.28316950
                    9   17.64038903
                    10  16.58264671
                    11  16.33172209
                    12  17.46773780
                };
                \addlegendentry{RoBERTa}
                
                \addplot+[mark=o, color=orange] table[x index=0, y index=1] {
                    1   8.30339491
                    2   9.86915474
                    3   12.04777235
                    4   13.84577020
                    5   15.53975246
                    6   16.27020456
                    7   17.06392695
                    8   17.72552642
                    9   17.18546907
                    10  16.18593110
                    11  16.01514044
                    12  17.17974739
                };
                \addlegendentry{RoBERTa fine-tuned on \MultiWOZ-train}
                
            \end{axis}
        \end{tikzpicture}
    \end{minipage}
    \caption{
        Comparison of the mean LID development through the different layers of {\RoBERTa}-base and a fine-tuned variant.
        The LID of embeddings from the distribution used for fine-tuning (here {\MultiWOZ}) differs significantly between the two models, whereas the LID of other data distributions (here {\Wikipedia}) is indistinguishable.
        \label{fig:twonn_fine_tuned_on_MultiWOZ_layerwise}
    }
\end{figure}

Our setup from \Cref{sec:methods} and \autoref{alg:compute_local_estimates} naturally applies to embeddings derived from arbitrary layers of the language model.
In particular, the mean local estimates can be compared layer-wise between different checkpoints of a model, as we show in \Cref{fig:twonn_fine_tuned_on_MultiWOZ_layerwise}.
The main observations here are that the drop in mean local dimension on the dataset used for fine-tuning is visible over all the model layers.
It appears to be most pronounced in the intermediate and last layers of the model.

\subsection{Additional Grokking Results}
\label{appendix:additional_grokking}

\begin{figure}[ht]
    \centering
        \begin{subfigure}{0.325\linewidth}
        \centering
        \includegraphics[width=1.0\linewidth]{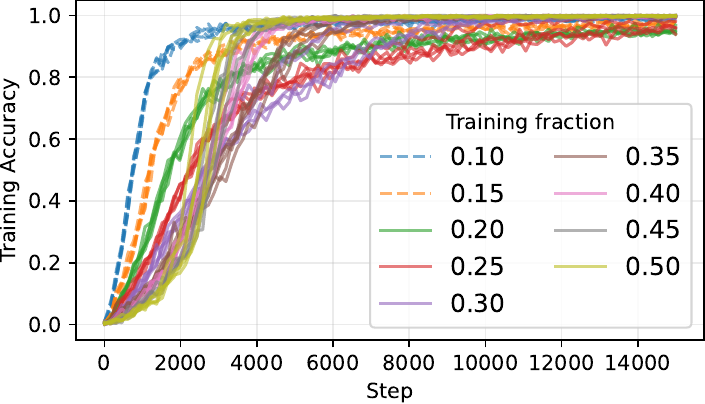}
        \caption{
            \textbf{Training} accuracy
            \label{fig:grokking_train.accuracy_individual_seeds}
        }
    \end{subfigure}
    \hfill
    \begin{subfigure}{0.325\linewidth}
        \centering
        \includegraphics[width=1.0\linewidth]{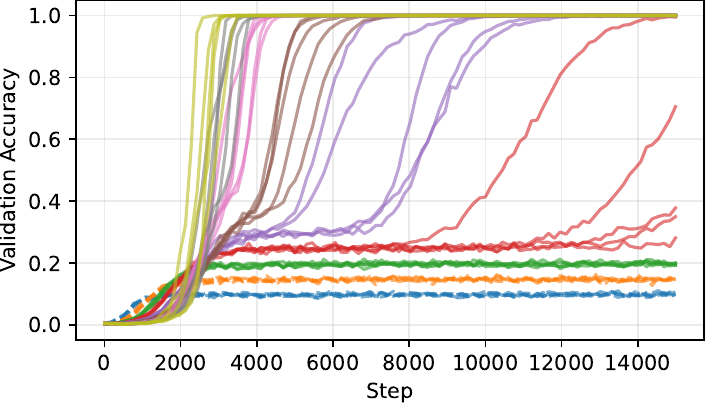}
        \caption{
            Validation accuracy
            \label{fig:grokking_val.accuracy_individual_seeds}
        }
    \end{subfigure}
    \hfill
    \begin{subfigure}{0.325\linewidth}
        \centering
        \includegraphics[width=1.0\linewidth]{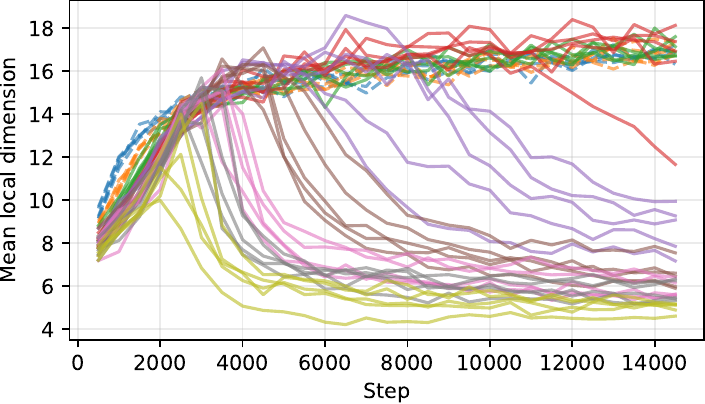}
        \caption{
            \textbf{Training} mean local estimates
            \label{fig:grokking_train.take_all.samples=3000.n-neighbors=64.mean_individual_seeds}
        }
    \end{subfigure}
    \caption{
        Training a model on addition mod \(p = 197\) with different training data fraction selected from \(\{0.1; 0.15; 0.2; 0.25; 0.3; 0.4; 0.5\}\).
        Five seeds per configuration are shown separately, demonstrating that the onset of grokking and the drop in mean local dimension vary greatly depending on the specific training run.
        The plots show the development over \num{15000} batches.
        The mean local estimates are computed on the \textbf{training split} for \(\TokenVectorSubsampleSize=3000\); \(\LocalNeighborhoodSize=64\).
        \label{fig:grokking_p_197_individual_seeds}
    }
\end{figure}

This section supplements the grokking experiments in \Cref{sec:local_dimensions_grokking}.
See \Cref{fig:grokking_p_197_individual_seeds} for a plot comparing model performance and mean local estimates on the training set, shown separately for the training run seeds.
This representation is useful, as the timing of the onset of grokking can be very dependent on the model and dataset shuffling seed.
It remains true that for individual seeds, the drop in mean local dimension on the training set coincides with the increase in validation accuracy.

%

\subsection{Additional TripPy-R Dialogue State Tracking Results}
\label{appendix:additional_trippy_r}

\begin{figure}[ht]
    \centering
    \begin{subfigure}{0.325\linewidth}
        \centering
        \includegraphics[width=\linewidth]{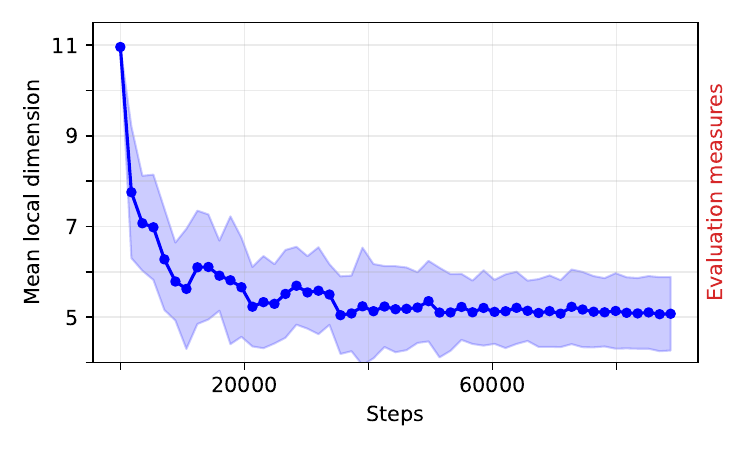}
        \caption{
            {\MultiWOZ} Training split
            \label{fig:trippy_r_multiwoz_50_epochs_linear_learning_rate_schedule_training}
        }
    \end{subfigure}
    \hfill
    \begin{subfigure}{0.325\linewidth}
        \centering
        \includegraphics[width=\linewidth]{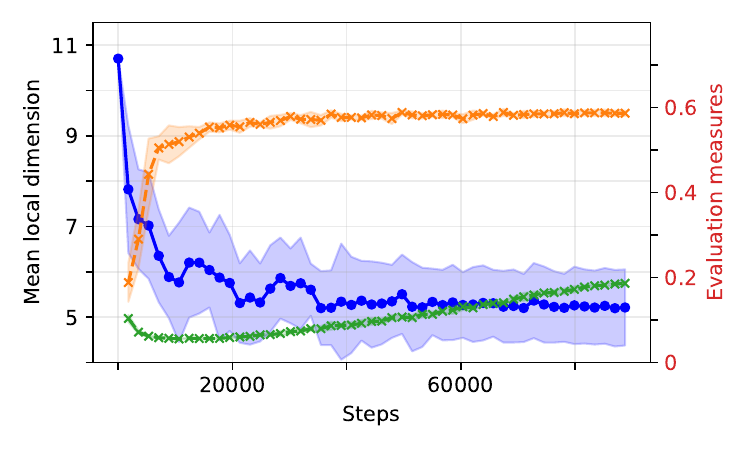}
        \caption{
            {\MultiWOZ} Validation split
            \label{fig:trippy_r_multiwoz_50_epochs_linear_learning_rate_schedule_validation}
        }
    \end{subfigure}
    \hfill
    \begin{subfigure}{0.325\linewidth}
        \centering
        \includegraphics[width=\linewidth]{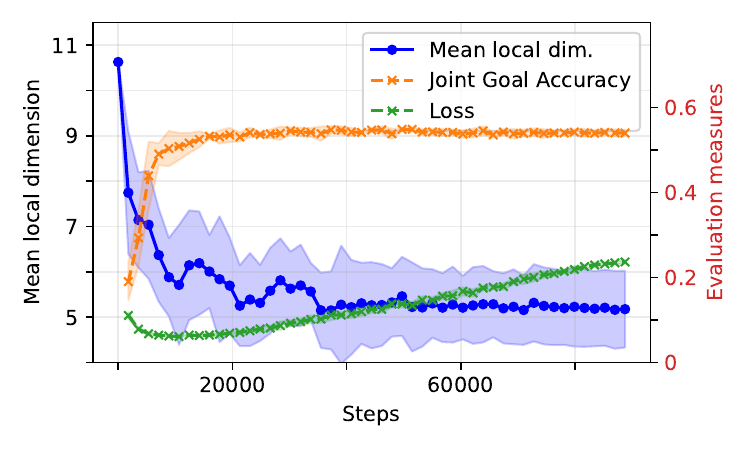}
        \caption{
            {\MultiWOZ} Test split
            \label{fig:trippy_r_multiwoz_50_epochs_linear_learning_rate_schedule_multiwoz_test}
        }
    \end{subfigure}
    \caption{
        Longer TripPy-R training runs over 50 epochs with a linear learning rate schedule.
        Shown are the mean and standard deviation of the measures over 4 seeds of the training runs.
        \label{fig:trippy_r_multiwoz_50_epochs_linear_learning_rate_schedule}
    }
\end{figure}

We here show additional results for \Cref{sec:local_dimensions_dialogue_state_tracking}, where we discuss exhausting model training capabilities.
See \Cref{fig:trippy_r_multiwoz_50_epochs_linear_learning_rate_schedule_multiwoz_test} for longer TripPy-R dialog state tracking runs over 50 epochs.
After about \(20\) training epochs, the mean local estimates stabilize and suggest the convergence of the model performance on the downstream task.

\clearpage

\end{document}